\documentclass[journal]{IEEEtran}
\usepackage[T1]{fontenc}
\usepackage{lscape}
\usepackage{tabularx}
\usepackage{array}
\usepackage{caption}
\usepackage{makecell}
\usepackage{mathtools} 
\usepackage{subcaption}  
\ifCLASSINFOpdf
   \usepackage[pdftex]{graphicx}
   \usepackage{epsfig}
\else
\fi
\usepackage{amsmath}
\usepackage[cmintegrals]{newtxmath}

\usepackage{booktabs}
\usepackage{multicol}
\usepackage{multirow}
\usepackage{titlesec}
\usepackage{url}
\usepackage{balance}

\let\oldtimes\times
\renewcommand{\times}{\mathord{\oldtimes}}

\renewcommand{\arraystretch}{1.2}

\begin{document}

\title{ProtoN: Prototype Node Graph Neural Network for Unconstrained Multi-Impression Ear Recognition}

\author{
    Santhoshkumar Peddi, Sadhvik Bathini, Arun Balasubramanian, Monalisa Sarma and Debasis~Samanta%
    \thanks{All authors are with the Department of Computer Science and Engineering, Indian Institute of Technology Kharagpur, Kharagpur, India.}%
    \thanks{Contact e-mails: santhoshpps11@gmail.com, sadhvik.ini@gmail.com, debasis.samanta.iitkgp@gmail.com}
}

\maketitle
\let\thefootnote\relax\footnotetext{This work has been submitted to the IEEE for possible publication. Copyright may be transferred without notice, after which this version may no longer be accessible.}

\begin{abstract}

Ear biometrics offer a stable and contactless modality for identity recognition, yet their effectiveness remains limited by the scarcity of annotated data and significant intra-class variability. Existing methods typically extract identity features from individual impressions in isolation, restricting their ability to capture consistent and discriminative representations. To overcome these limitations, a few-shot learning framework, \textit{ProtoN}, is proposed to jointly process multiple impressions of an identity using a graph-based approach. Each impression is represented as a node in a class-specific graph, alongside a learnable prototype node that encodes identity-level information. This graph is processed by a Prototype Graph Neural Network (PGNN) layer, specifically designed to refine both impression and prototype representations through a dual-path message-passing mechanism. To further enhance discriminative power, the PGNN incorporates a cross-graph prototype alignment strategy that improves class separability by enforcing intra-class compactness while maintaining inter-class distinction. Additionally, a hybrid loss function is employed to balance episodic and global classification objectives, thereby improving the overall structure of the embedding space. Extensive experiments on five benchmark ear datasets demonstrate that ProtoN achieves state-of-the-art performance, with Rank-1 identification accuracy of up to 99.60\% and an Equal Error Rate (EER) as low as 0.025, showing the effectiveness for few-shot ear recognition under limited data conditions.

\end{abstract}


\IEEEpeerreviewmaketitle

\section{Introduction}

\IEEEPARstart{B}{iometric} authentication systems have increasingly sought reliable and stable physiological characteristics for accurate human identification. Among various biometric modalities, the human ear presents exceptional advantages due to its structural permanence and distinctive features that remain consistent throughout an individual's lifetime (from eight to seventy years old) \cite{ahila2021deep, iannarelli1964}. Its inherent geometric complexity offers rich discriminative information while maintaining resilience to aging effects, facial expressions, and minor injuries \cite{iannarelli1964}. Additionally, ear-based systems enable contactless authentication through distant image capture, eliminating the need for physical interaction and making them particularly suitable for security applications where hygiene and user convenience are paramount.

Despite these compelling advantages, ear recognition has not achieved widespread deployment, mainly due to limitations in the existing recognition frameworks. A predominant approach in the literature involves adapting convolutional neural networks (CNNs) pre-trained on large-scale visual datasets to ear recognition tasks using transfer learning \cite{benzaoui2023comprehensive, el2022exploring, chowdhury2022privacy, emervsic2023unconstrained}. However, these general-purpose architectures often fail to capture the fine-grained geometric and textural cues specific to ear structures, resulting in suboptimal performance \cite{emervsivc2019unconstrained, kornblith2019better}. Alternatively, training models from scratch is constrained by the scarcity of annotated ear datasets \cite{emersic2017ear, kinear2022, emervsic2023unconstrained, kumar2012automated, amiear}, which restricts their ability to generalize across diverse subjects and conditions. Beyond data scarcity, a more fundamental limitation lies in how feature representations are constructed within existing recognition pipelines. Most approaches \cite{benzaoui2023comprehensive, emervsic2023unconstrained} treat ear images independently, processing each impression in isolation without considering whether multiple impressions of the same identity are available. Even when datasets contain varied poses, lighting conditions, or occlusions for the same subject, these impressions are passed through the network separately. Information fusion occurs only after feature extraction, often through simplistic methods such as averaging or score-level combination \cite{mesquita2020rethinking}, which disregards the rich contextual relationships between impressions and fails to recover identity-specific cues that emerge only through collective modeling of multiple impressions \cite{HE2021107930}.

This work proposes a paradigm shift toward structured multi-impression modeling during feature extraction to address these limitations. Rather than treating each image as an independent sample, the proposed approach jointly models subsets of impressions as structured entities through a graph-based formulation. Graph neural networks (GNNs) are naturally suited for this task as they excel at modeling relational dependencies between entities through learnable message passing mechanisms \cite{rong2020deep, han2022vision}. In this framework, impressions are represented as nodes with their relationships encoded through message passing, allowing the network to integrate both local impression features and global identity context dynamically.

However, while enhancing representation richness, this graph-based formulation simultaneously exacerbates the data scarcity problem by reducing the number of effective training instances per identity. To overcome this challenge, the proposed method is embedded within a prototypical few-shot learning framework that enables generalization across unseen identities using only a small number of support examples \cite{snell2017prototypical}. This integration provides a distance-based classification mechanism in an embedding space, ensuring that each prototype is formed through structured aggregation of multiple impressions, collected across several graphs, via the graph-based encoder.

This paper introduces \textit{ProtoN}, a graph-based few-shot learning architecture for ear recognition that simultaneously addresses the challenges of representation quality and data scarcity. It represents each identity using multiple impression graphs, enabling relational modeling of intra-class variation through message passing. A cross-graph prototype alignment strategy ensures consistency across impression subsets of the same class. At the same time, a hybrid loss formulation mitigates embedding space crowding \cite{laenen2021episodes, zou2019hierarchical} and enhances discriminability under few-shot conditions.

The key contributions of this work are as follows:
\begin{itemize}
    \item A graph-based architecture for structured multi-impression modeling that captures inter-impression dependencies during feature extraction.
    \item A learnable prototype node embedded within each graph to serve as a unified identity representation.
    \item A cross-graph alignment mechanism that enhances intra-class compactness and inter-class separation.
    \item A hybrid loss function that balances episodic training with global embedding structure for improved few-shot classification.
    \item A novel integration of multi-impression feature extraction and few-shot learning for ear biometrics, offering improved generalization from limited annotated data.
\end{itemize}

The remainder of this paper is organized as follows. Section 2 reviews related work in ear recognition and few-shot learning. Section 3 presents the proposed methodology. Section 4 discusses experimental results and analysis. Section 5 concludes the paper and outlines future research directions.

\section{Related Work}
The field of ear recognition has its historical roots in the pioneering work of Bertillon and McClaughry, who first recognised that the shape of the ear could be utilised for criminal identification \cite{bertillon1896}. Because ear images are in limited supply, the predominant strategy has been to employ transfer learning as demonstrated by Emeršič \textit{et al.} \cite{emervsicdataagu} through the use of SqueezeNet architectures and by Chowdhury \textit{et al.} \cite{chowdhury2022privacy} using DenseNet 161 models for privacy-preserving ear recognition. Building upon this foundation, similar fine-tuning methodologies have been extended by El-Naggar and Bourlai \cite{el2022exploring}, where DenseNet models were adapted for both visible and thermal ear image recognition.

Beyond individual model fine-tuning, ensemble learning has emerged as a powerful technique that enhances recognition performance. This approach was initially explored by Eyiokur \textit{et al.} \cite{eyiokur2018domain} through the fusion of VGG16 and GoogLeNet models based on confidence scores. The effectiveness of ensemble methods subsequently inspired Alshazly \textit{et al.} \cite{alshazly2021towards} to combine ResNet101 and ResNet152 architectures, where predictions were averaged to improve overall accuracy. More recently, this ensemble paradigm has been further refined by Kumar and Agarwal \cite{kumar2023unconstrained} through integrating EfficientNet and ViT models using triplet loss and late fusion strategies for unconstrained ear recognition. Similar approaches using traditional machine learning classifiers include the work by, Sharkas \cite{sharkas2022ear}, who used ResNet50 as a feature extractor coupled with a subspace discriminant ensemble classifier and Mehta \textit{et al.} \cite{mehta2024ensemble} employed VGG16 and VGG19 networks as feature extractors, feeding the extracted features separately into SVM classifiers to obtain prediction scores, which were then averaged to produce the final prediction.

Developing novel deep learning architectures designed explicitly for ear recognition has also gained considerable attention. A notable contribution was introduced by Priyadarshini \textit{et al.} \cite{ahila2021deep}, who developed a custom Deep CNN model tailored for ear image identification. This architecture was subsequently enhanced by Alomari \textit{et al.} \cite{alomari2024ear} through integrating pix2pix GAN technology for data augmentation purposes. Innovation in this domain continued with Korichi \textit{et al.} \cite{korichi2022tr}, who introduced a CNN-based model incorporating tied rank (TR) normalization applied to ICA-processed CNN filters. This work employed a multi-modal approach combining TR-ICANet, TR-PCANet, AlexNet, and VGG19 feature extractors with SVM classifiers for final prediction. Following similar principles, Aiadi \textit{et al.} \cite{aiadi2023mdfnet} modified the PCANet architecture to create MDFNet, which utilizes SVM classifiers for unsupervised lightweight ear recognition.

\begin{figure*}[!t]
    \centering    
    \includegraphics[width=\textwidth]{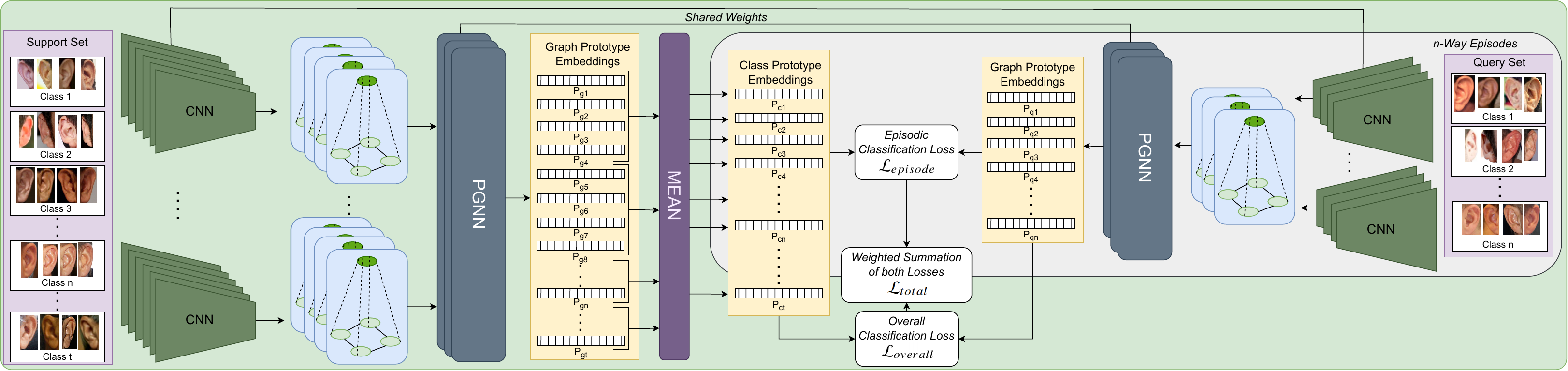}
    \caption{Overview of the graph-based prototypical network architecture that extracts class prototype and graph prototype embeddings from support and query sets through shared CNN and PGNN layers, respectively, with distance computation enabling few-shot classification under a dual-mode training objective.}
    \label{overview}
\end{figure*}

Recent advances have explored specialized network architectures and attention mechanisms for improved ear recognition performance. A Siamese network approach incorporating Squeeze-Excitation Attention Module (SE-SiamNet) with ResNet50 as the backbone has been demonstrated by He \textit{et al.} \cite{he2024self} with promising results. Additionally, an innovative pipeline employing CLIPAsso has been investigated by Freire-Obregón \textit{et al.} \cite{freire2025synthesizing} to extract sketches from ear images, followed by recognition using DenseNet121 and a triplet loss.

In order to advance research in ear recognition under real-world, unconstrained conditions, the research community has organized several unconstrained ear recognition challenges (UERC) to foster innovation and provide standardized evaluation frameworks. The inaugural challenges organized by Emeršič \textit{et al.} in 2017 \cite{emervsivc2017unconstrained} and 2019 \cite{emervsivc2019unconstrained} introduced the UERC dataset series and stimulated the development of numerous innovative ear recognition methods through competitive evaluation. The most recent UERC 2023 challenge \cite{emervsic2023unconstrained} yielded several state-of-the-art approaches, including the Deep HOG-CNN Fusion (DHCF) method, the KU-EAR system utilizing a pretrained ResNet18 backbone, and the PreWAdaEar approach based on fine-tuned AdaFace models. Additional notable contributions from this challenge include the MEM-Ear multi-algorithm ensemble approach, the UERC-IGD using a baseline ResNet18-based model for multitask learning using CosFace Loss, the RecogEAR system employing a two-stream inflated 3D ConvNet, and the ViTEar method that concatenated the embeddings from the three fine-tuned DINOv2 networks using margin penalty softmax losses.

Although the methods described above have made significant contributions to advancing ear biometrics research, they suffer from limitations related to generalizability, suboptimal recognition rates, and a lack of dedicated ear detection systems for practical deployment. These challenges have motivated the development of a novel approach, which is presented comprehensively in the subsequent section.

\section{Methodology}
This section outlines ProtoN, a graph-based few-shot learning framework for ear recognition that models multiple impressions of an identity using graph representations. Each graph consists of impression nodes and a learnable prototype node, which collectively encode local and global identity information through stacked Prototype Graph Neural Network (PGNN) layers. As shown in Fig. \ref{overview}, support and query images are passed through a shared CNN to extract embeddings, which are then organized into multiple graphs per identity. Within the support set, the prototype node from each graph serves as a feature vector, and their average forms the class prototype. The refined graph prototype is used directly for distance-based classification in the query set. This structure enables the network to capture inter-impression relationships, promote discriminative class representations, and generalize well under limited supervision. The following subsections describe the graph construction, PGNN message passing, and hybrid training strategy in detail.

\subsection{Graph Construction}
To model intra-class variation while preserving identity-specific structure, each class is represented as a collection of impression-level graphs. For every class, \( K \) graphs are constructed, each composed of \( N \) ear images, resulting in \( M = K \times N \) impressions per class. Each graph is formally defined as \( G_g = (V_g, E_g) \), where \( V_g \) denotes the set of nodes and \( E_g \subseteq V_g \times V_g \) specifies the undirected edges connecting them.

The node set \( V_g \) consists of \( N \) real nodes \( h_i \), each corresponding to an individual ear image, along with a single prototype node \( p_g \) representing the aggregated identity context. Each image \( x_i \in \mathbb{R}^{H \times W \times 3} \), where \( i = 1, \dots, N \), is passed through a shared convolutional encoder to generate a feature embedding \( h_i^{(0)} \in \mathbb{R}^d \), as defined in (\ref{eq:embedding}):

\begin{equation}
h_i^{(0)} = \text{AvgPool2D}\left( \sigma\left( \text{MaxPool}(\text{BN}(\text{Conv}(x_i))) \right) \right)_4
\label{eq:embedding}
\end{equation}

where \( \sigma(\cdot) \) denotes the ReLU activation function, \( \text{BN}(\cdot) \) denotes batch normalization, and the subscript 4 indicates the use of four sequential convolutional blocks. The resulting vector \( h_i^{(0)} \) serves as the initial representation of the \( i \)-th real node. This embedding network is trained end-to-end, allowing the CNN to learn ear-specific features tailored to the task, rather than depending on generic pre-trained backbones.

Once the real node embeddings are computed, the prototype node \( p_g^{(0)} \in \mathbb{R}^d \) is initialized as shown in (\ref{eq:prototype}):

\begin{equation}
p_g^{(0)} = \frac{1}{N} \sum_{i=1}^{N} h_i^{(0)}
\label{eq:prototype}
\end{equation}

This prototype acts as a central representation of the class and enables interaction between the individual impressions and a shared identity abstraction.

The edge set \( E_g \) governs how information propagates through the graph. It includes two types of connections. The first links each real node to its immediate neighbors using modulo indexing, forming a cyclic structure that establishes edges of the form \( (v_i, v_{i+1}) \) and \( (v_i, v_{i-1}) \). This configuration promotes localized message passing while maintaining graph sparsity and balanced connectivity. Its minimal edge design avoids unnecessary complexity while ensuring sufficient context exchange across impressions. The second set of connections links each real node directly to the prototype node, forming bidirectional edges \( (v_i, p_g) \) that allow impression-level features to influence and be influenced by the identity-level representation. The complete edge set is defined in (\ref{eq:edges}):

\begin{equation}
E_g = \{(v_i, v_j) \mid j = (i \pm 1) \bmod N\} \cup \{(v_i, p_g) \mid 1 \leq i \leq N\}
\label{eq:edges}
\end{equation}

The full connectivity is encoded in a binary adjacency matrix \( A \in \mathbb{R}^{(N+1) \times (N+1)} \), where the final row and column correspond to the prototype node. The adjacency entries are defined as given in (\ref{eq:adjacency}):

\begin{equation}
a_{ij} = 
\begin{cases}
1, & \text{if } j = (i + 1) \bmod N \text{ or } j = (i - 1) \bmod N \\
1, & \text{if } j = p_g \\
0, & \text{otherwise}
\end{cases}
\label{eq:adjacency}
\end{equation}

This hybrid topology captures both local inter-impression dependencies and global identity context within a compact, computation-friendly structure. These constructed graphs are subsequently passed to the PGNN layers, where iterative message passing refines impression-specific and prototype-level representations.

\subsection{PGNN Message Passing}
Refinement of node embeddings in ProtoN is achieved through stacked PGNN layers, which apply dedicated update rules to real and prototype nodes. These updates allow the network to model both impression-level variation and identity-level structure by controlling how information is aggregated within and across graphs.

\subsubsection{Real Node Update}
The update of a real node in each PGNN layer begins (see (\ref{realnode1}))  with a non-linear transformation of the node’s features:
\begin{equation}
h_i^{(l+1)} = \sigma \left( W_r^{(l)} h_i^{(l)} \right)
\label{realnode1}
\end{equation}
where \( h_i^{(l)} \) is the feature of node \( i \) at layer \( l \), \( W_r^{(l)} \) is a learnable transformation matrix, and \( \sigma(\cdot) \) is the ReLU activation. This operation enables the node to refine its internal representation before interacting with other nodes.

Following this self-update, the node aggregates information from its neighbors \( j \in N(i) \) to enrich its understanding of surrounding relationships. Instead of treating all neighbors equally, the model employs a learned attention mechanism that assigns dynamic weights to each connection, as shown in (\ref{realnode2}):

\begin{equation}
h_i^{(l+1)} = \sigma \left( 
W_r^{(l)} h_i^{(l)} +
\sum_{j \in N(i)} \alpha_{ij}^{(l)} U_r^{(l)} h_j^{(l)}
\right)
\label{realnode2}
\end{equation}

Here, \( U_r^{(l)} \) applies a feature-wise transformation to each neighbor, while \( \alpha_{ij}^{(l)} \) denotes the attention weight assigned to node \( j \). These attention scores are computed based on the similarity between the transformed embeddings of the source and target nodes, as defined in (\ref{eq_attention}):
\begin{equation}
\alpha_{ij}^{(l)} = 
\frac{
\exp \left( \left( W_{\alpha}^{(l)} h_i^{(l)} \right)^\top \left( W_{\alpha}^{(l)} h_j^{(l)} \right) \right)
}{
\sum_{k \in N(i)} \exp \left( \left( W_{\alpha}^{(l)} h_i^{(l)} \right)^\top \left( W_{\alpha}^{(l)} h_k^{(l)} \right) \right)
}
\label{eq_attention}
\end{equation}

Where \( W_{\alpha}^{(l)} \) is a projection matrix that maps the original feature into a common attention space. 

While neighborhood-level context captures intra-graph variation, it lacks global identity awareness. To bridge this gap, each node additionally receives guidance from the prototype node \( p_g^{(l)} \) of its graph \( g \). This prototype aggregates features across all impressions in the graph and, through interaction with other prototypes (detailed next), encodes rich identity-level context.

The node adjusts its feature toward this prototype using the difference \( p_g^{(l)} - h_i^{(l)} \), which acts as a correction signal pointing toward the semantic center of the graph. The strength of this adjustment is modulated by a learned coefficient \( \beta_i^{(l)} \), allowing flexible influence based on the node’s current state. The full update rule is described in (\ref{realnode}):

\begin{equation}
\begin{split}
h_i^{(l+1)} = \sigma \bigg( 
& W_r^{(l)} h_i^{(l)} + \sum_{j \in N(i)} \alpha_{ij}^{(l)} U_r^{(l)} h_j^{(l)} \\
& + \beta_i^{(l)} U_p^{(l)} \left( p_g^{(l)} - h_i^{(l)} \right)
\bigg)
\end{split}
\label{realnode}
\end{equation}

The matrix \( U_p^{(l)} \) transforms the prototype correction term, while \( \beta_i^{(l)} \) is a scalar gate predicting how much the prototype should influence node \( i \). This gating value is derived from the node’s current feature using (\ref{eq_beta}):
\begin{equation}
\beta_i^{(l)} = \sigma \left( W_{\beta}^{(l)} h_i^{(l)} \right)
\label{eq_beta}
\end{equation}

where \( W_{\beta}^{(l)} \) is a learnable projection that maps the feature to a scalar in the range \( (0, 1) \), which allows the model to apply stronger correction to outlier or under-refined nodes, while preserving the representation of those already aligned with the prototype.

\subsubsection{Prototype Node Update}

The prototype node \( p_g \) is integrated into each graph as a representative anchor that encodes impression-level diversity and higher-level identity context. To serve this dual role effectively, the prototype must capture intra-class variation, ensuring it reflects a class's diverse impressions while maintaining inter-class separation to support discriminative classification. This balance is particularly critical in few-shot settings, where class boundaries are learned from limited data. The prototype is progressively refined through self-transformation, feedback from real nodes, and alignment with other prototypes across the episode to achieve this.

The update begins with a self-transformation that stabilizes the representation of the prototype as it evolves through the layers, as shown in (\ref{protonode1}):
\begin{equation}
p_g^{(l+1)} = \sigma \left( W_p^{(l)} p_g^{(l)} \right)
\label{protonode1}
\end{equation}
Here, \( W_p^{(l)} \) is a learnable transformation matrix and \( \sigma(\cdot) \) is the ReLU activation. This step retains previously learned information while preparing the prototype for contextual refinement.
To ensure the prototype remains reflective of its associated impressions, a feedback mechanism is introduced where each real node \( h_i^{(l)} \) contributes a residual update. These updates are modulated by adaptive coefficients \( \gamma_i^{(l)} \), which determine the reliability of each node in influencing the prototype. The updated formulation becomes as given in (\ref{protonode2}):

\begin{equation}
p_g^{(l+1)} = \sigma \left( W_p^{(l)} p_g^{(l)} + \sum_{i=1}^{N} \gamma_i^{(l)} U_r'^{(l)} (h_i^{(l)} - p_g^{(l)}) \right)
\label{protonode2}
\end{equation}

where \( U_r'^{(l)} \) is a learnable transformation matrix for real-node feedback. The influence weights \( \gamma_i^{(l)} \) are computed using the current prototype representation as described in (\ref{eq_gamma}):

\begin{equation}
\gamma_i^{(l)} = \sigma \left( W_\gamma^{(l)} p_g^{(l)} \right)
\label{eq_gamma}
\end{equation}

This feedback step ensures the prototype remains grounded in impression-level features and captures the intra-class variation present within the graph.

To complement this local refinement, a cross-graph alignment term is incorporated to expose the prototype to a broader class-level context. This term allows \( p_g \) to communicate with all other prototypes \( p_{g'} \) in the episode, enabling it to align with semantically similar graphs and differentiate from dissimilar ones. The complete update formulation is presented in (\ref{protonode}):

\begin{equation}
\begin{split}
p_g^{(l+1)} = \sigma \Big(
W_p^{(l)} p_g^{(l)} +
\sum_{i=1}^{N} \gamma_i^{(l)} U_r'^{(l)} (h_i^{(l)} - p_g^{(l)}) \\
+ \lambda \sum_{g' \neq g} w_{gg'} U_p'^{(l)} (p_{g'}^{(l)} - p_g^{(l)})
\Big)
\end{split}
\label{protonode}
\end{equation}

In this equation, \( \lambda \) controls the influence of the alignment, \( U_p'^{(l)} \) is a learnable transformation matrix for prototype-level interaction, and \( w_{gg'} \in (0,1) \) is a similarity weight indicating how strongly graph \( g' \) should affect graph \( g \). These weights are computed as (see (\ref{eq_cross_graph})).

\begin{equation}
w_{gg'} = \sigma \left( W_w^{(l)} [p_g^{(l)} \, || \, p_{g'}^{(l)}] \right)
\label{eq_cross_graph}
\end{equation}

where \( [\cdot \, || \, \cdot] \) denotes feature concatenation and \( W_w^{(l)} \) is a learnable projection matrix. This alignment mechanism improves semantic coherence within a class and enhances inter-class discrimination by encouraging separation between unrelated prototypes.

The second term in (\ref{protonode}) denotes the feedback from real nodes, which helps capture impression-level variation, while the third term (cross-graph alignment) structures the prototypes in a way that promotes inter-class separability. This joint formulation enables each prototype to act as a refined summary of its own graph and a semantically informed representation within the broader class space.

The prototypes in the query set are restricted from interacting with other graphs to ensure they generalize independently to unseen classes. Instead, they are updated using only the self-transformation and internal feedback terms, as specified in (\ref{qprotonode}):

\begin{equation}
p_g^{(l+1)} = \sigma \left( W_p^{(l)} p_g^{(l)} + \sum_{i=1}^{N} \gamma_i^{(l)} U_r'^{(l)} (h_i^{(l)} - p_g^{(l)}) \right)
\label{qprotonode}
\end{equation}

This structure maintains the purity of few-shot generalization and is empirically supported by ablation results, which show a performance decline when cross-graph alignment is applied to query prototypes.

\subsection{Hybrid Prototypical Training}
After $l$ layers, the refined prototype $p_g$ for each graph in support $s_i$ is collected. The class prototype $p_c$ is computed as the average of all graph prototypes belonging to that class as given in (\ref{eq:class_prototype}).

\begin{equation}
\label{eq:class_prototype}
p_c = \frac{1}{|G_c|} \sum_{g \in G_c} p_g
\end{equation}

where $G_c$ is the set of graphs for class $c$, and $p_g$ is the prototype for graph $g$.

Similarly, a refined query prototype $p_q$ is computed for the query graphs and is classified using the softmax over negative distances across classes $c_i$ as given in (\ref{eq:classification_prob}).

\begin{equation}
\label{eq:classification_prob}
P(y=c|q) = \frac{\exp(-d(q,p_c))}{\sum_{c'} \exp(-d(q,p_{c'}))}
\end{equation}

where $d(\cdot,\cdot)$ is the Euclidean distance function. A negative log-likelihood loss function is selected in (\ref{eq:loss}) to provide smooth gradients.

\begin{equation}
\label{eq:loss}
\mathcal{L} = -\log P(y=c|q) = -\log \frac{\exp(-d(q,p_c))}{\sum_{c'} \exp(-d(q,p_{c'}))}
\end{equation}

where $\mathcal{L}$ represents the cross-entropy loss computed over a softmax of negative distances, measuring the negative log-likelihood of assigning the query $q$ to its true class $c$.

\begin{figure}[!t]
    \centering
    \includegraphics[width=0.5\textwidth]{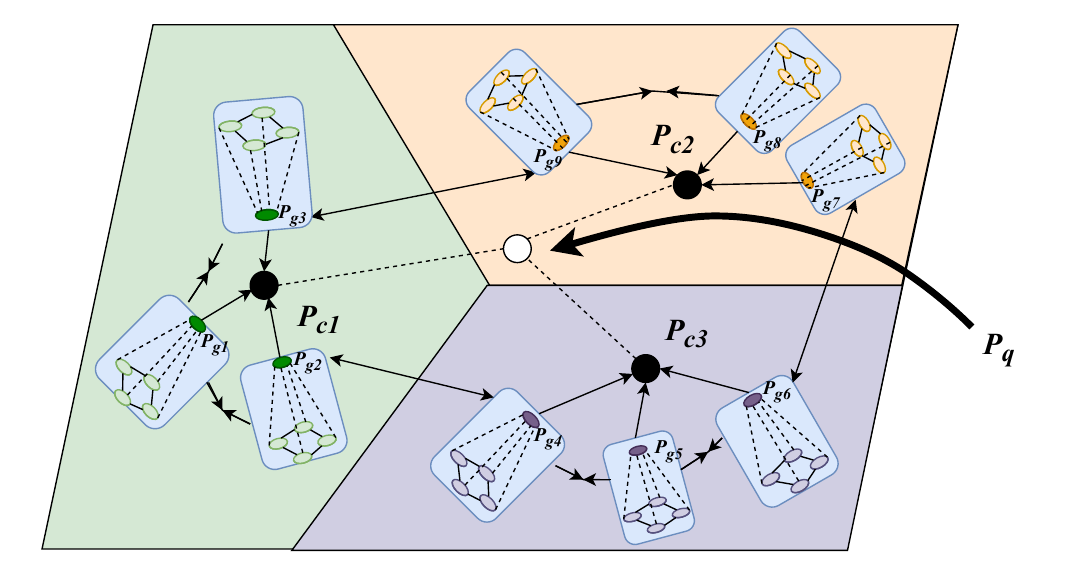}
    \caption{Interaction of class prototypes, graphs' prototype nodes and query prototype in embedding space}
    \label{space}
\end{figure}

It must be noted that optimising solely within each episode made the embedding space crowded, causing it to become particularly problematic because class representations would collapse into similar regions as the number of ways increased (see Fig. \ref{space}). This led to the identification of a critical limitation in episode-based training. To overcome this challenge, the model must know more classes during each episode to create more separable embeddings \cite{snell2017prototypical}. However, increasing the ways in each episode is computationally complex and unreliable. Hence, in every episode, a weighted hybrid loss (see (\ref{eq:hybrid_loss})) is used, which helps the model to learn discriminative features.

\begin{equation}
\label{eq:hybrid_loss}
\mathcal{L}_{total} = (1 - \lambda) \cdot \mathcal{L}_{episode} + \lambda \cdot \mathcal{L}_{overall}
\end{equation}

where $\lambda \in [0,1]$ is a weighting parameter that controls the trade-off between the episode-specific loss $\mathcal{L}_{episode}$ and the overall loss $\mathcal{L}_{overall}$. The optimal value of $\lambda$ may be chosen empirically (see Section \ref{sssec: lambda in Hybrid Loss}).

\section{Experiments \& Analysis}

This section presents a structured evaluation of the proposed method, guided by the following objectives:
\begin{itemize}
    \item Train the ProtoN using a few-shot classification setup and evaluate its ability to learn reliable identity representations from multiple impressions per class.
    \item Demonstrate the model’s effectiveness through recognition tasks, including identification and verification, across various datasets.
    \item Compare the proposed method with existing approaches to highlight improvements in accuracy, generalization, and overall performance.
    \item Conduct ablation studies to analyze the role of each architectural component and training strategy in the overall performance.
\end{itemize}

\subsection{Datasets}
To ensure a comprehensive evaluation, experiments were conducted across five publicly available ear recognition datasets that vary in scale, image conditions, and subject diversity:

\begin{enumerate}
    \item Unconstrained Ear Recognition Challenge (UERC-2023) \cite{emervsic2023unconstrained}: A large-scale dataset combining UERC 2017/2019, Annotated Web Ears (AWE), AWEx, CVL, and additional web-crawled sources. It includes 1310 identities with 10–500 impressions per identity.

    \item Annotated Web Ears (AWE) \cite{emersic2017ear}: Collected from web images of public figures, this dataset includes 100 classes with 10 unconstrained impressions per class. It is also part of UERC-2023.

    \item Indian Institute of Technology Delhi – Version II (IITD-II) \cite{kumar2012automated}: Captured indoors using a touchless setup, it contains 3–5 ear images each from 221 subjects.

    \item Mathematical Analysis of Images (AMI) \cite{amiear}: Comprises 700 high-resolution images from 100 individuals (7 per class), acquired in controlled lab settings.

    \item Kinship Ear (KinEar) \cite{kinear2022}: Initially designed for kinship verification, it includes 76 identities with 15–30 unconstrained, web-sourced images per class. Portions of it are also used in UERC-2023.
\end{enumerate}

\subsection{Training and Evaluation}
ProtoN was trained using a prototypical learning framework designed to handle few-shot ear biometric recognition. The UERC-2023 dataset classes were randomly partitioned into 70\% training, 15\% testing, and 15\% validation splits, where the training and testing splits were used for model training purposes, and the validation split was kept aside for recognition experiments. As the UERC-2023 dataset already contains the AWE dataset images, those classes with 10 images were excluded from the training split by assuming that they belong to the AWE dataset, and the model was retrained with the modified UERC dataset and tested. An episodic training strategy was employed using 5-way 4-Graphs and 10-way 4-Graphs configurations, with episodes randomly sampled within the training split with replacement. Data augmentation techniques were applied to the training split to balance impressions across classes, following established practices for addressing class imbalance in biometric datasets \cite{emervsicdataagu}. The model consists of a 4-layer CNN feature extractor followed by three PGNN layers. Training was conducted on NVIDIA A40 GPUs with hyperparameters detailed in Table \ref{hyperparameters}.

\begin{table}[!t]
\caption{Hyperparameters}
\label{hyperparameters}
\centering
\begin{tabular}{l | l}
\hline
\textbf{Parameter} & \textbf{Value} \\
\hline
Resized Image Dimensions & 128 X 128 \\
Image Normalization (Mean) & [0.485, 0.456, 0.406]\\
Image Normalization (Std) & [0.229, 0.224, 0.225]\\
Total Number of CNN Layers & 4 \\
Number of Channels (Each Layer) in CNN & 64, 128, 256, 512\\
CNN Feature vector Dimension ($h^{(0)}$) & 512 \\
Total Number of PN-GNN Layers ($l$) & 3\\
Number of Input Channels in PN-GNN & 512 \\
Number of Hidden Channels in PN-GNN & 256 \\
Number of Output Channels in PN-GNN & 128 \\
Input \& Output Dimensions of MLP & 128, 128\\
Training Episodes per epoch & 200 \\
Testing episodes & 100 \\
Number of Graphs ($K$) & 4\\
Number of Images (with replacement) ($N$)
& 5\\
Total number of Learnable parameters & 3183628 \\
Learning Rate & 0.001\\
Number of Training Epochs & 1000 \\
Optimizer & Adam \\
Loss Adjustment Weight ($\lambda$) & 0.4\\
Random Seed & 42\\
\hline
\end{tabular}
\end{table}

To evaluate the model's generalization capacity, the trained model was tested across multiple benchmark datasets with and without fine-tuning. Performance evaluation was conducted using two distinct metrics: episodic and overall accuracy. Episodic accuracy measures the model's performance within individual few-shot learning episodes, calculated as the percentage of correctly classified query samples within each episode, then averaged across all episodes. Overall accuracy simultaneously evaluates the model's performance across all classes in the dataset, providing a comprehensive measure of cross-class discrimination capability. The initial evaluation used the frozen model trained solely on UERC-2023 to assess direct transfer capabilities, with results presented in Table \ref{results_without_finetuning}. Subsequently, fine-tuning experiments were conducted on each target dataset, where 70\% of the classes were used for training and the remaining 30\% reserved for evaluation, as shown in Table~\ref{results_with_finetuning}.

\begin{figure*}[!t]
    \centering
    \begin{subfigure}[b]{0.49\linewidth}
        \centering
        \includegraphics[width=\linewidth]{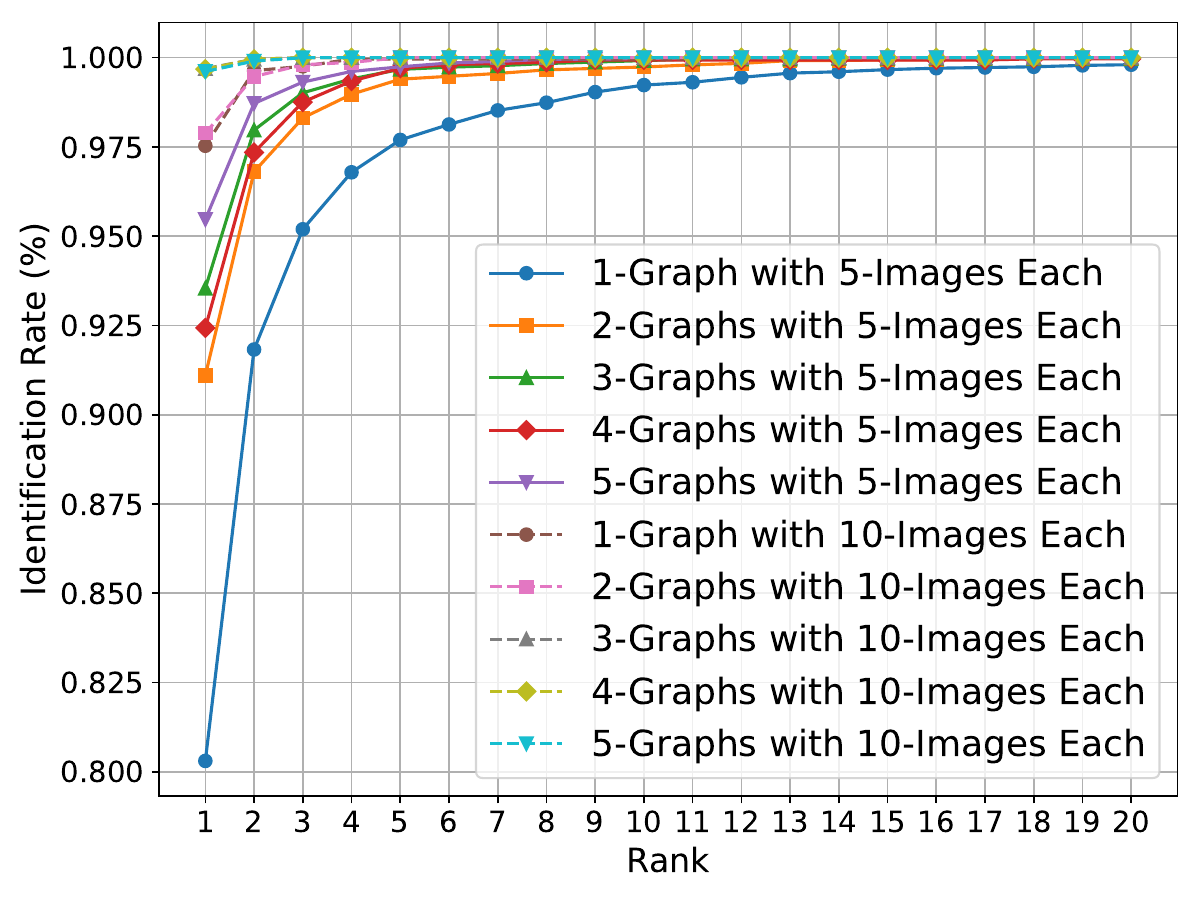}
        \caption{}
        \label{fig:UERC-cmc}
    \end{subfigure}
    \hfill
    \begin{subfigure}[b]{0.48\linewidth}
        \centering
        \includegraphics[width=\linewidth]{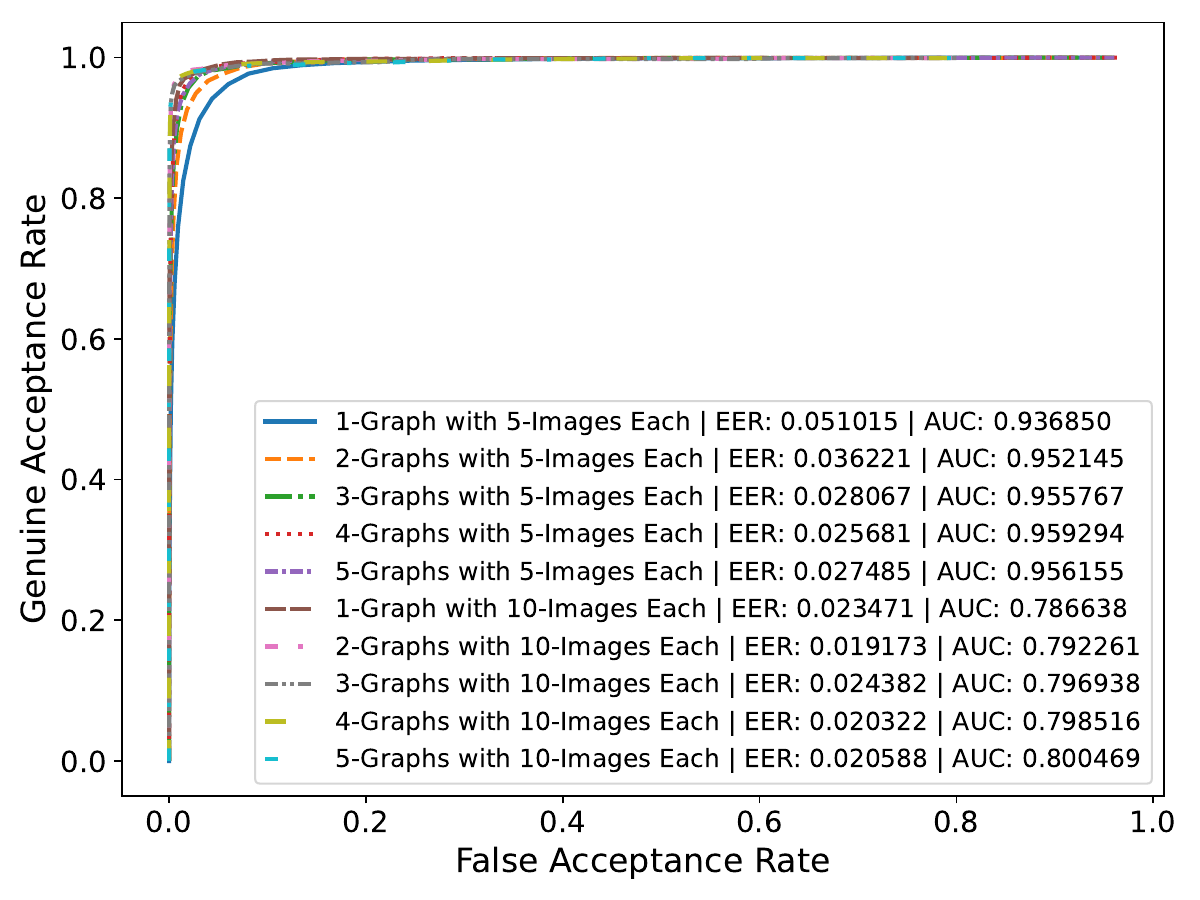}
        \caption{}
        \label{fig:UERC-roc}
    \end{subfigure}
    \caption{Recognition results on the UERC dataset across prototype configurations: (a) CMC curves for identification, (b) ROC curves for verification.}
    \label{fig:uerc-recognition}
\end{figure*}

\begin{table}[!t]
\centering
\caption{Few-shot learning performance trained on UERC and tested across different datasets}
\label{results_without_finetuning}
\renewcommand{\arraystretch}{1.5}
\begin{tabular}{@{}p{1cm}|l|cc|cc@{}}
\hline
\multirow{2}{*}{\textbf{Acc. (\%)}} & \multirow{2}{*}{\textbf{Dataset}} & \multicolumn{2}{c|}{\textbf{5-Way}} & \multicolumn{2}{c}{\textbf{10-Way}} \\
\cline{3-6}
& & \textbf{1-Graph} & \textbf{4-Graphs} & \textbf{1-Graph} & \textbf{4-Graphs} \\
\hline
\multirow{5}{*}{\rotatebox{90}{\small\textbf{\begin{tabular}{c}Episodic\\Accuracy\end{tabular}}}} 
& UERC & 92.20 & \textbf{99.84} & 90.70 & \textbf{99.36} \\
& AWE & 92.80 & 99.99 & 94.30 & 99.65 \\
& IITD-II & 97.00 & 99.89 & 96.80 & 99.42 \\
& AMI & 94.00 & 98.50 & 80.00 & 94.10 \\
& KinEar & 76.00 & 93.20 & 62.50 & 88.82 \\
\hline
\multirow{5}{*}{\rotatebox{90}{\small\textbf{\begin{tabular}{c}Overall\\Accuracy\end{tabular}}}} 
& UERC & 49.79 & \textbf{88.89} & 64.59 & \textbf{88.97} \\
& AWE & 78.59 & 92.84 & 88.99 & 94.72 \\
& IITD-II & 55.59 & 81.09 & 63.59 & 80.57 \\
& AMI & 73.99 & 71.59 & 60.09 & 70.67 \\
& KinEar & 47.19 & 68.54 & 45.99 & 69.12 \\
\hline
\end{tabular}
\end{table}

The comparison between these two evaluation scenarios reveals the model's inherent capacity for cross-dataset generalization, with minimal performance differences between frozen and fine-tuned versions across most datasets. This demonstrates that the learned representations capture fundamental ear biometric features that transfer effectively across different data distributions, making the approach suitable for practical biometric recognition applications where labeled target domain data may be limited.

\begin{table}[!t]
\centering
\caption{Few-shot learning performance across different datasets with Fine-tuning}
\label{results_with_finetuning}
\renewcommand{\arraystretch}{1.5}  
\begin{tabular}{@{}p{1cm}|l|cc|cc@{}}
\hline
\multirow{2}{*}{\textbf{Acc. (\%)}} & \multirow{2}{*}{\textbf{Dataset}} & \multicolumn{2}{c|}{\textbf{5-Way}} & \multicolumn{2}{c}{\textbf{10-Way}} \\
\cline{3-6}
& & \textbf{1-Graph} & \textbf{4-Graphs} & \textbf{1-Graph} & \textbf{4-Graphs} \\
\hline
\multirow{4}{*}{\rotatebox{90}{\small\textbf{\begin{tabular}{c}Episodic\\Accuracy\end{tabular}}}} 
& AWE & 90.20 & 99.34 & 93.00 & 99.12 \\
& IITD-II & 99.60 & 99.94 & 97.80 & 99.92 \\
& AMI & 99.80 & 99.84 & 98.90 & 99.95 \\
& KinEar & 86.00 & 98.89 & 87.80 & 94.10 \\
\hline
\multirow{4}{*}{\rotatebox{90}{\small\textbf{\begin{tabular}{c}Overall\\Accuracy\end{tabular}}}} 
& AWE & 98.39 & 97.04 & 98.29 & 97.12 \\
& IITD-II & 96.19 & 95.35 & 95.29 & 94.07 \\
& AMI & 99.98 & 98.34 & 99.89 & 99.79 \\
& KinEar & 87.39 & 98.24 & 96.89 & 94.79 \\
\hline
\end{tabular}
\end{table}

\subsection{Recognition Results}
To evaluate the practical applicability of ProtoN, comprehensive recognition experiments were conducted encompassing both identification and verification tasks. The experimental design centers around testing multiple prototype configurations, where each configuration is defined by K graphs with N images per graph. Specifically, configurations with  $K \in \{1, 2, 3, 4, 5\}$ and $N \in \{5, 10\}$ were evaluated, resulting in combinations such as "4 graphs with 5 images", "2 graphs with 10 images". Each configuration represents a different strategy for computing enrollment prototypes, where multiple graphs containing varying numbers of images are aggregated to form robust identity representations. Furthermore, the experiments were conducted across all datasets using frozen and fine-tuned model versions. 

\subsubsection{Identification Results}

The identification experiments evaluate the model's ability to identify individuals from a closed-set gallery of enrolled prototypes correctly. Each query prototype is matched against all enrolled prototypes to determine the most likely identity match. Performance is measured using Cumulative Match Characteristic (CMC) curves that illustrate the probability of finding the correct match within the top-k ranked candidates.

This experiment's dataset partitioning was carefully designed to form distinct enrollment and test sets. For classes with sufficient samples, where the total number of impressions \( r \) significantly exceeds the required number \( K \times N \), the first \( K \times N \) impressions are assigned for enrollment, and the remaining ones are reserved for testing. In cases where \( r \leq K \times N \), approximately half of the available impressions (\( \lfloor r/2 \rfloor \)) are allocated to the enrollment set and the remainder to the test set, with replacement performed independently within each set to meet configuration requirements. Importantly, impressions used for enrollment and testing are kept strictly disjoint. While the same impressions may appear across different graphs within the same set (enrollment or test), each graph is constructed using unique impressions. If the number of available impressions is insufficient even to form a single complete graph, additional samples are synthetically generated by injecting controlled noise into existing impressions to meet the structural constraints.

Figure~\ref{fig:UERC-cmc} shows the corresponding identification performance, which presents CMC curves for multiple prototype configurations on the primary dataset. Extended plots for all remaining datasets, including both fine-tuned and frozen settings, are provided in Fig. \ref{fig:s1} and \ref{fig:s2}. In those results, fine-tuning consistently improves Rank-1 accuracy across conditions, while non-fine-tuned models maintain stable performance, reflecting effective generalization.

\begin{table*}[!t]
\caption{Comparing the proposed method using Rank Accuracies (trained on UERC Dataset) with the existing works with \& without finetuning on other datasets. Values in \textbf{bold} indicate the top score for that column.}
\label{table:compare_rank}
\centering
\begin{tabular}{l|c|c|c|c|c|c|c|c|c|c}
\hline
\multicolumn{1}{c|}{\textbf{Method}} & \multicolumn{5}{c|}{\textbf{Rank-1 Accuracy (\%) $\uparrow$}} & \multicolumn{5}{c}{\textbf{Rank-5 Accuracy (\%) $\uparrow$}} \\
\cline{2-11}
& \textbf{UERC} & \textbf{AWE} & \textbf{IITD-II} & \textbf{AMI} & \textbf{KinEar} & \textbf{UERC} & \textbf{AWE} & \textbf{IITD-II} & \textbf{AMI} & \textbf{KinEar} \\
\hline
\cite{emervsicdataagu} & 62.00 & -- & -- & -- & -- & 80.35  & -- & -- & -- & -- \\
\cite{dodge2018unconstrained} & -- & 68.50 & -- & -- & -- & -- & 85.00 & -- & -- & -- \\
\cite{eyiokur2018domain} & 6.9 & -- & -- & -- & -- & -- & -- & -- & -- & -- \\
ExplainableEar\cite{alshazly2021towards} & -- & 67.25 & -- & \textbf{99.64} & -- & -- & 85.50 & -- & \textbf{100} & -- \\
\cite{el2022exploring} & -- & 98.915 & -- & -- & -- & -- & -- & -- & -- & -- \\
\cite{chowdhury2022privacy} & -- & 50.50 & -- & -- & -- & -- & 70.00 & -- & -- & -- \\
MDFNet\cite{aiadi2023mdfnet} & -- & 82.5 & \textbf{98.96} & 97.67 & -- & -- & -- & -- & -- & -- \\
\cite{kumar2023unconstrained} & -- & 41.49 & -- & -- & 57.69 & -- & 65.99 & -- & -- & 78.74 \\
ViTEar\cite{emervsic2023unconstrained} & 96.27 & -- & -- & -- & -- & -- & -- & -- & -- & -- \\
Pix2Pix-GAN\cite{alomari2024ear} & -- & -- & -- & 98.00 & -- & -- & -- & -- & -- & -- \\
Ensemble\cite{mehta2024ensemble} & -- & -- & 29.00 & -- & -- & -- & -- & 93.00 & -- & -- \\
EarSketch\cite{freire2025synthesizing} & -- & 20.3 & 20.0 & 24.2 & -- & -- & 46.2 & 51.4 & 64.7 & -- \\
\hline
\textbf{ProtoN} & \textbf{99.60} & \textbf{100} & 70.00 & 55.00 & 55.14 & \textbf{100} & \textbf{100} & 91.00 & 79.00 & 80.51 \\
\textbf{ProtoN (Fine-Tuned)} & \textbf{99.60} & \textbf{100} & 85.07 & 96.66 & \textbf{90.00} & \textbf{100} & \textbf{100} & \textbf{94.02} & \textbf{100} & \textbf{100} \\
\hline
\end{tabular}
\end{table*}

\begin{table*}[!t]
\caption{Comparing the proposed method using EER (trained on UERC Dataset) with the existing works with \& without finetuning on other datasets. Values in \textbf{bold} indicate the top score for that column.}
\label{compare_eer}
\centering
\begin{tabular}{l|c|c|c|c|c|c|c|c|c|c}
\hline
\multicolumn{1}{c|}{\textbf{Method}} & \multicolumn{5}{c|}{\textbf{EER $\downarrow$}} & \multicolumn{5}{c}{\textbf{AUC $\uparrow$}} \\
\cline{2-11}
& \textbf{UERC} & \textbf{AWE} & \textbf{IITD-II} & \textbf{AMI} & \textbf{KinEar} & \textbf{UERC} & \textbf{AWE} & \textbf{IITD-II} & \textbf{AMI} & \textbf{KinEar} \\
\hline
ExplainableEar\cite{alshazly2021towards} & -- & -- & -- & -- & -- & -- & 0.960 & -- & \textbf{0.989} & -- \\
\cite{el2022exploring} & -- & 0.0075 & -- & -- & -- & -- & -- & -- & -- & -- \\
Ensemble\cite{mehta2024ensemble} & -- & -- & -- & -- & -- & -- & -- & 0.56 & -- & -- \\
MEM-EAR\cite{emervsic2023unconstrained} & 0.146 & -- & -- & -- & -- & 0.915 & -- & -- & -- & -- \\
ViTEar\cite{emervsic2023unconstrained} & 0.177 & -- & -- & -- & -- & 0.908 & -- & -- & -- & -- \\
DHCF\cite{emervsic2023unconstrained} & 0.185 & -- & -- & -- & -- &0.895 & -- & -- & -- & -- \\
IGD\cite{emervsic2023unconstrained} & 0.190 & -- & -- & -- & -- &0.868 & -- & -- & -- & -- \\
KU-EAR\cite{emervsic2023unconstrained} & 0.198 & -- & -- & -- & -- &0.880 & -- & -- & -- & -- \\
PreWAdaEAR\cite{emervsic2023unconstrained} & 0.204 & -- & -- & -- & -- &0.887 & -- & -- & -- & -- \\
RecogEAR\cite{emervsic2023unconstrained} & 0.493 & -- & -- & -- & -- &0.494 & -- & -- & -- & -- \\
UERC Baseline\cite{emervsic2023unconstrained} & 0.360 & -- & -- & -- & -- & 0.699 & -- & -- & -- & -- \\
\hline
\textbf{ProtoN} & \textbf{0.025} & 0.009 & 0.092 & 0.103 & 0.195 & \textbf{0.959} & 0.827 & 0.737 & 0.917 & 0.867 \\
\textbf{ProtoN (Fine-Tuned)} & \textbf{0.025} &  \textbf{0.0005} &  \textbf{0.085} &  \textbf{0.013} & \textbf{0.077} & \textbf{0.959} &  \textbf{0.999} & \textbf{0.975} & 0.981 & \textbf{0.963} \\
\hline
\end{tabular}
\end{table*}

\subsubsection{Verification Results}
For the verification task, the evaluation is conducted by comparing pairs of prototypes to determine whether they belong to the same identity (genuine) or different identities (imposter). In all configurations, one of the prototypes in each pair is fixed as a single graph constructed from $N$ impressions \( (1 \times N) \), representing a consistent and limited test-time input. The other prototype varies based on the chosen configuration and is constructed using $K$ graphs with $N$ impressions per graph (e.g., $1 \times 10$, $2 \times 5$, $3 \times 10$, etc.), allowing analysis of how richer representations affect verification performance.

Genuine pairs are formed by sampling a $1 \times N $ prototype and a $K \times N$ prototype from the same identity using disjoint subsets of impressions. If the number of available impressions is insufficient, sampling with replacement is applied while ensuring no overlap between the two prototypes in a pair. Imposter pairs are formed by selecting a $1 \times N $ prototype from one identity and a  $K \times N$ prototype from a different identity, following the same construction rules. This setup ensures consistent evaluation across varying levels of prototype complexity while maintaining structural balance in each comparison.

Verification is performed by computing Euclidean distances between prototype pairs, with performance evaluated using ROC curves and Equal Error Rate (EER) as the primary metric. Figure~\ref{fig:UERC-roc} shows the result on the primary dataset, while Fig. \ref{fig:s3} and \ref{fig:s4} present extended results on additional datasets under frozen and fine-tuned settings. Fine-tuned models achieve lower EERs overall, while frozen versions retain strong generalization. A similar pattern can be observed with the AUC values obtained for the ROC plots as shown in Table \ref{compare_eer}.

\subsection{Comparison with the Existing Methods}
The proposed method is evaluated on identification and verification tasks across five benchmark datasets. All comparative results from existing methods are taken directly from their respective papers. Performance is assessed under two settings: direct generalization using the model trained on UERC, and fine-tuning on each target dataset.

Table~\ref{table:compare_rank} summarizes the identification results regarding Rank-1 and Rank-5 accuracy. The proposed method consistently achieves top performance across datasets, even without fine-tuning. On datasets such as AWE and AMI, generalization performance is already competitive with specialized models, reflecting strong domain transfer capabilities. After fine-tuning, the model outperforms all baselines, achieving near-saturation performance in most cases. Notably, the gain in Rank accuracy highlights the model’s ability to learn discriminative and structurally consistent prototypes, even under high inter-class similarity.

Table~\ref{compare_eer} presents verification results in terms of EER and AUC. The model demonstrates low error rates and high discriminative capability across all datasets. Without fine-tuning, performance on UERC already surpasses many existing methods, suggesting effective representation learning during training. After fine-tuning, verification accuracy improves significantly on target datasets. The drop in EER on AWE and AMI shows that the model effectively adapts to new distributions while maintaining robust identity separation. These results suggest that the graph-based prototype aggregation and hybrid loss formulation contribute directly to improved generalization and calibration in the verification task compared to baselines.

Overall, the results highlight the robustness and adaptability of the proposed method. It consistently outperforms or matches state-of-the-art approaches in both identification and verification, under both generalization and fine-tuned settings, confirming its effectiveness for few-shot ear recognition across diverse real-world datasets.

\subsection{Ablation Study}
A comprehensive ablation study was conducted to validate the effectiveness of individual components within the proposed architecture. Each experiment systematically removed or altered a specific module while keeping all other configurations fixed, enabling isolation of its contribution. All ablation variants were trained for 250 epochs on the UERC dataset for consistency, ensuring a fair and controlled comparison.

\subsubsection{Impact of Multi-Impression Graph Modeling}
The model was restructured to operate with only a single impression per class during training and inference to evaluate the necessity of multi-impression aggregation and graph-based reasoning. As a result, the graph construction and message-passing mechanisms were removed, and the model processed isolated embeddings without relational context. All remaining components were kept unchanged, including the hybrid loss formulation and prototypical learning objective.

This variant consistently underperforms during training due to the absence of relational context across impressions. Recognition metrics also show a clear drop in performance and an increase in error rate, as reported in Table~\ref{abrecog}, highlighting the importance of multi-impression graph modeling for learning robust and identity-consistent representations. The corresponding accuracy plot is shown in Fig. \ref{fig:s5}.

\subsubsection{Impact of Cross-Graph Prototype Alignment}
The cross-graph alignment term defined in (\ref{protonode}) enforces consistency among prototypes of the same identity across different graphs. This term was removed in the ablation variant to isolate its effect on the training and recognition processes. All other components, including prototype node interactions and the hybrid loss, remained active to ensure a controlled comparison.

While the test accuracy during training remains similar to the original model, the recognition performance degrades, especially in Rank-1 accuracy and EER (Table~\ref{abrecog}). These results indicate that cross-graph alignment does not influence convergence directly but is essential for producing prototypes that generalize well to unseen samples. Fig. \ref{fig:s5} shows the comparison of the training progression.

\subsubsection{Impact of Prototype Alignment in Query Graphs} \label{sssec: Query graph communication}
Although cross-graph alignment improves prototype consistency during training, it is deliberately excluded from query graph processing to ensure independent inference. An ablation variant was constructed where the alignment term was also applied during query updates, introducing inter-query dependencies that conflict with the few-shot inference setting, where each query should be processed in isolation to reflect real-time deployment.

Test accuracy curves remain similar, but recognition performance suffers when alignment is applied during inference, as seen in Table~\ref{abrecog}, confirming that preserving query-level independence is necessary for generalization and reliable few-shot inference. The accuracy plot comparison is available in Fig. \ref{fig:s5}.

\subsubsection{Role of the Prototype Node in the PGNN Layer}
The prototype node within the PGNN layer serves as a learnable class-level representation that interacts with impression nodes through message passing. It aggregates information from the graph, refines node features, and aligns with other prototypes across graphs to enhance representation consistency.

The prototype node and its connections were removed to assess its role, and the graph-level representation was instead derived via global average pooling over impression nodes. This change resulted in slower convergence and a noticeable drop in recognition performance, as evidenced by Table~\ref{abrecog}, confirming the importance of the prototype node. A comparative plot illustrating this effect is shown in Fig. \ref{fig:s5}.

\subsubsection{Influence of Hybrid Loss Weighting Parameter$(\lambda)$} \label{sssec: lambda in Hybrid Loss}

To enhance prototype discrimination and mitigate class overlap in the embedding space, the proposed method incorporates a hybrid loss function (see (\ref{eq:hybrid_loss})) that combines episodic and overall class-level objectives. The two components are balanced using a scalar weight \(\lambda\), where \(\lambda = 0.0\) corresponds to using only the overall loss \(L_{\text{all}}\), and \(\lambda = 1.0\) uses only the episodic loss \(L_{\text{episode}}\).

\begin{figure}
    \centering
    \includegraphics[width=\linewidth]{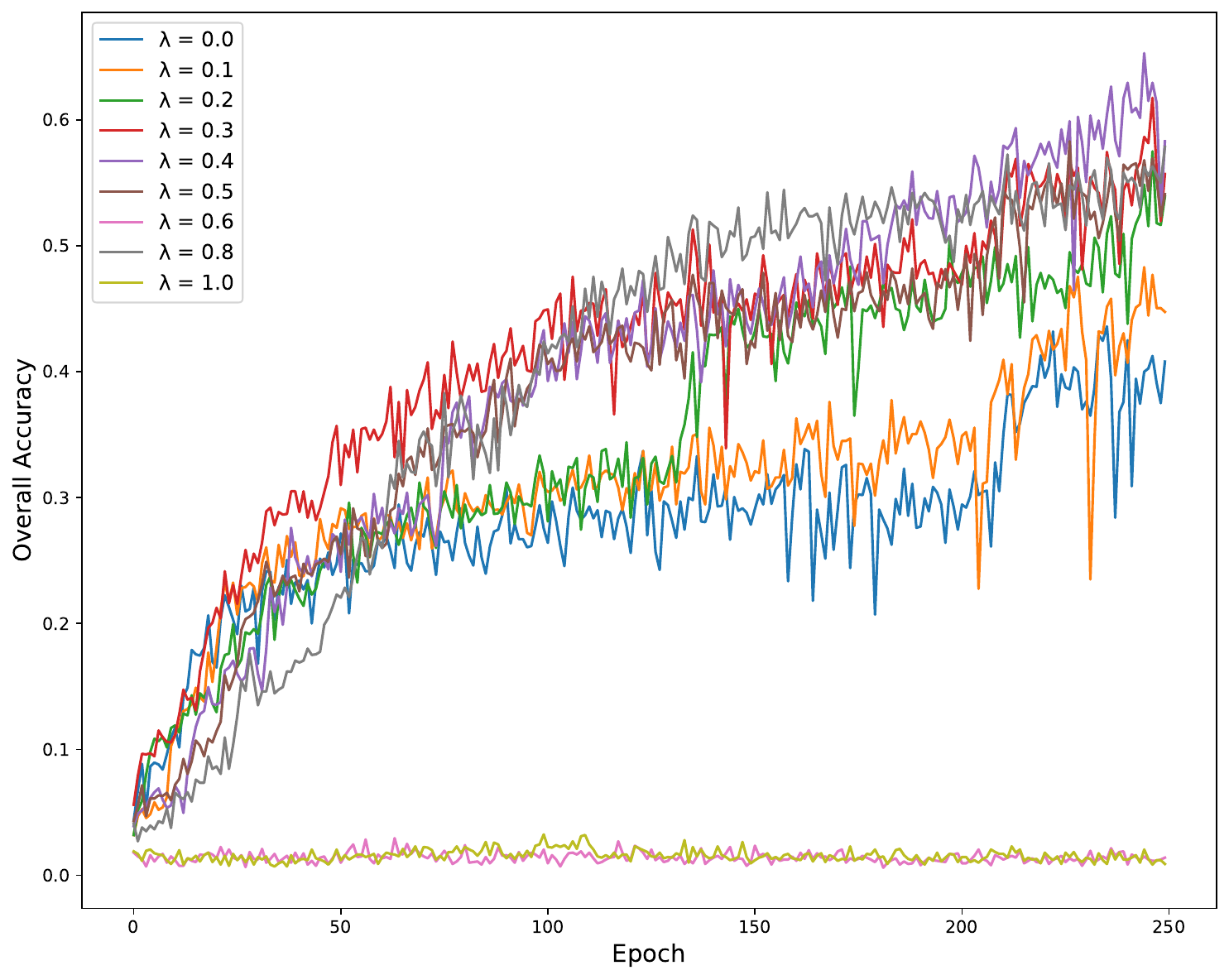}
    \caption{Effect of Hybrid Loss Weight $\lambda$}
    \label{fig:placeholder}
\end{figure}

The model was trained to identify the optimal balance with different values of \(\lambda\) ranging from 0.0 to 1.0. The results in Fig. \ref{fig:placeholder} indicate that the best overall accuracy is achieved at \(\lambda = 0.4\), supporting hybrid supervision. Performance trends related to \(\lambda\) selection are consistent with those reported in Table~\ref{abrecog}.

\subsubsection{Ablation Summary}
\begin{table}[!t]
\renewcommand{\arraystretch}{1.5}
\caption{Recognition results during ablation study on UERC dataset.}
\label{abrecog}
\centering
\begin{tabularx}{\linewidth}{>{\raggedright}p{3.8cm}|X|X|p{0.9cm}}
\hline
\textbf{Ablation Setting} & \textbf{Rank-1 $\uparrow$} & \textbf{Rank-5 $\uparrow$} & \textbf{EER $\downarrow$} \\
\hline
Impact of Multi-Impression Graph Modeling & 4.01 & 17.67 & 0.1520 \\
Impact of Cross-Graph Prototype Alignment & 64.71 & 93.81 & 0.0568 \\
Impact of Prototype Alignment in Query Graphs & 62.35 & 91.81 & 0.0596 \\
Role of the Prototype Node in the PGNN Layer & 20.01 & 53.46 & 0.1639 \\
\hline
Proposed Method (ProtoN) & 91.97 & 99.64 & 0.025 \\
\hline
\end{tabularx}
\end{table}

To consolidate the findings from all ablation settings, Table~\ref{abrecog} summarizes the recognition performance of each variant. It highlights how individual components contribute to improved recognition accuracy. The complete model consistently outperforms ablated versions, confirming the effectiveness of each design choice.

\section{Conclusions}
This paper introduced \textit{ProtoN}, a graph-based few-shot learning approach tailored for ear biometric recognition, emphasizing structured multi-impression modeling and prototype alignment. Experiments demonstrated substantial improvements in recognition performance, notably achieving increases of up to 15–20\% in Rank-1 accuracy and significant reductions in EER compared to existing CNN-based methods. Such performance gains highlight the potential of leveraging relational information between multiple impressions to enhance robustness, particularly in data-constrained biometric applications.

However, \textit{ProtoN}'s current implementation necessitates multiple impressions per identity at inference, posing practical limitations for scenarios with single-image availability. Additionally, the method's effectiveness partly depends on dataset richness, as indicated by improved prototype stability with increased impressions per class and graphs per identity. Future work may explore strategies to support single-impression inference and improve robustness under more constrained settings, contributing to broader applicability in real-world ear recognition systems.

\balance
\bibliographystyle{IEEEtran}
\bibliography{references}

\begin{thebibliography}{10}
\providecommand{\url}[1]{#1}
\csname url@samestyle\endcsname
\providecommand{\newblock}{\relax}
\providecommand{\bibinfo}[2]{#2}
\providecommand{\BIBentrySTDinterwordspacing}{\spaceskip=0pt\relax}
\providecommand{\BIBentryALTinterwordstretchfactor}{4}
\providecommand{\BIBentryALTinterwordspacing}{\spaceskip=\fontdimen2\font plus
\BIBentryALTinterwordstretchfactor\fontdimen3\font minus
  \fontdimen4\font\relax}
\providecommand{\BIBforeignlanguage}[2]{{%
\expandafter\ifx\csname l@#1\endcsname\relax
\typeout{** WARNING: IEEEtran.bst: No hyphenation pattern has been}%
\typeout{** loaded for the language `#1'. Using the pattern for}%
\typeout{** the default language instead.}%
\else
\language=\csname l@#1\endcsname
\fi
#2}}
\providecommand{\BIBdecl}{\relax}
\BIBdecl

\bibitem{ahila2021deep}
R.~Ahila~Priyadharshini, S.~Arivazhagan, and M.~Arun, ``A deep learning
  approach for person identification using ear biometrics,'' \emph{Appl.
  intell.}, vol.~51, no.~4, pp. 2161--72, 2021.

\bibitem{iannarelli1964}
A.~V. Iannarelli, \emph{The Iannarelli System of Ear Identification}.\hskip 1em
  plus 0.5em minus 0.4em\relax Foundation Press, 1964.

\bibitem{benzaoui2023comprehensive}
A.~Benzaoui, Y.~Khaldi, R.~Bouaouina, N.~Amrouni, H.~Alshazly, and A.~Ouahabi,
  ``A comprehensive survey on ear recognition: databases, approaches,
  comparative analysis, and open challenges,'' \emph{Neurocomput.}, vol. 537,
  pp. 236--70, 2023.

\bibitem{el2022exploring}
S.~El-Naggar and T.~Bourlai, ``Exploring deep learning ear recognition in
  thermal images,'' \emph{IEEE Trans. on Biometrics, Behav., and Identity
  Sci.}, vol.~5, no.~1, pp. 64--75, 2022.

\bibitem{chowdhury2022privacy}
D.~P. Chowdhury, S.~Bakshi, C.~Pero, G.~Olague, and P.~K. Sa, ``Privacy
  preserving ear recognition system using transfer learning in industry 4.0,''
  \emph{IEEE Trans. on Ind. Inform.}, vol.~19, no.~5, pp. 6408--17, 2022.

\bibitem{emervsic2023unconstrained}
{\v{Z}}.~Emer{\v{s}}i{\'c}, T.~Ohki, M.~Akasaka, T.~Arakawa, S.~Maeda,
  M.~Okano, Y.~Sato, A.~George, S.~Marcel, I.~I. Ganapathi \emph{et~al.}, ``The
  unconstrained ear recognition challenge 2023: Maximizing performance and
  minimizing bias,'' in \emph{Int. Jt. Conf. on Biometrics (IJCB)}, 2023, pp.
  1--10.

\bibitem{emervsivc2019unconstrained}
{\v{Z}}.~Emer{\v{s}}i{\v{c}}, A.~K. S.~V., B.~Harish, W.~Gutfeter, J.~Khiarak,
  A.~Pacut, E.~Hansley, M.~P. Segundo, S.~Sarkar, H.~Park \emph{et~al.}, ``The
  unconstrained ear recognition challenge 2019,'' in \emph{Int. Conf. on
  Biometrics (ICB)}, 2019, pp. 1--15.

\bibitem{kornblith2019better}
S.~Kornblith, J.~Shlens, and Q.~V. Le, ``Do better imagenet models transfer
  better?'' in \emph{Proceedings of the IEEE/CVF conference on computer vision
  and pattern recognition}, 2019, pp. 2661--2671.

\bibitem{emersic2017ear}
{\v{Z}}.~Emer{\v{s}}i{\v{c}}, V.~{\v{S}}truc, and P.~Peer, ``Ear recognition:
  More than a survey,'' \emph{Neurocomput.}, vol. 255, pp. 26--39, 2017.

\bibitem{kinear2022}
G.~Dvor{\v{s}}ak, A.~Diwedi, V.~{\v{S}}truc, P.~Peer, and
  {\v{Z}}.~Emer{\v{s}}i{\v{c}}, ``Kinship verification from ear images: An
  explorative study with deep learning models.''\hskip 1em plus 0.5em minus
  0.4em\relax IEEE, 2022, pp. 1--6.

\bibitem{kumar2012automated}
A.~Kumar and C.~Wu, ``Automated human identification using ear imaging,''
  \emph{Pattern Recognit.}, vol.~45, no.~3, pp. 956--68, 2012.

\bibitem{amiear}
G.~Esther, A.~Luis, and M.~Luis, ``Ami ear database,''
  \url{https://webctim.ulpgc.es/research_works/ami_ear_database/}, accessed 01
  August 2025.

\bibitem{mesquita2020rethinking}
D.~Mesquita, A.~H. Souza, and S.~Kaski, ``Rethinking pooling in graph neural
  networks,'' in \emph{Proc. of the Int. Conf. on Neural Inf. Process. Syst.
  (NIPS)}.\hskip 1em plus 0.5em minus 0.4em\relax Curran Associates Inc., 2020,
  pp. 2220--31.

\bibitem{HE2021107930}
H.~He and S.~Chen, ``Identification of facial expression using a multiple
  impression feedback recognition model,'' \emph{Appl. Soft Comput.}, vol. 113,
  p. 107930, 2021.

\bibitem{rong2020deep}
Y.~Rong, T.~Xu, J.~Huang, W.~Huang, H.~Cheng, Y.~Ma, Y.~Wang, T.~Derr, L.~Wu,
  and T.~Ma, ``Deep graph learning: Foundations, advances and applications,''
  in \emph{Proc. of the ACM SIGKDD int. conf. on knowl. discov. \& data min.},
  2020, pp. 3555--6.

\bibitem{han2022vision}
K.~Han, Y.~Wang, J.~Guo, Y.~Tang, and E.~Wu, ``Vision gnn: an image is worth
  graph of nodes,'' in \emph{Proc. of the Int. Conf. on Neural Inf. Process.
  Syst. (NIPS)}.\hskip 1em plus 0.5em minus 0.4em\relax Curran Associates Inc.,
  2022, pp. 8291--303.

\bibitem{snell2017prototypical}
J.~Snell, K.~Swersky, and R.~Zemel, ``Prototypical networks for few-shot
  learning,'' in \emph{Proc. of the Int. Conf. on Neural Inf. Process. Syst.
  (NIPS)}.\hskip 1em plus 0.5em minus 0.4em\relax Curran Associates Inc., 2017,
  pp. 4080--90.

\bibitem{laenen2021episodes}
S.~Laenen and L.~Bertinetto, ``On episodes, prototypical networks, and few-shot
  learning,'' \emph{Adv. in Neural Inf. Process. Syst.}, vol.~34, pp.
  24\,581--92, 2021.

\bibitem{zou2019hierarchical}
Y.~Zou and J.~Feng, ``Hierarchical meta learning,'' 2019.

\bibitem{bertillon1896}
A.~Bertillon and R.~W. McClaughry, \emph{Signaletic Instructions Including the
  Theory and Practice of Anthropometrical Identification}.\hskip 1em plus 0.5em
  minus 0.4em\relax Werner Company, 1896.

\bibitem{emervsicdataagu}
{\v{Z}}.~Emer{\v{s}}i{\'c}, D.~{\v{S}}tepec, V.~{\v{S}}truc, and P.~Peer,
  ``Training convolutional neural networks with limited training data for ear
  recognition in the wild,'' in \emph{Int. Conf. on Autom. Face and Gesture
  Recognit. (FG)}, 2017, pp. 987--94.

\bibitem{eyiokur2018domain}
F.~I. Eyiokur, D.~Yaman, and H.~K. Ekenel, ``Domain adaptation for ear
  recognition using deep convolutional neural networks,'' \emph{IET
  Biometrics}, vol.~7, no.~3, pp. 199--206, 2018.

\bibitem{alshazly2021towards}
H.~Alshazly, C.~Linse, E.~Barth, S.~A. Idris, and T.~Martinetz, ``Towards
  explainable ear recognition systems using deep residual networks,''
  \emph{IEEE Access}, vol.~9, pp. 122\,254--73, 2021.

\bibitem{kumar2023unconstrained}
V.~Kumar and A.~Agarwal, ``On unconstrained ear recognition for
  privacy-preserving authentication,'' in \emph{Wksp. on Adv. of Mob. and
  Wearable Biometric (WAMWB)}, 2023, pp. 21--36.

\bibitem{sharkas2022ear}
M.~Sharkas, ``Ear recognition with ensemble classifiers; a deep learning
  approach,'' \emph{Multimed. Tools and Appl.}, vol.~81, no.~30, pp.
  43\,919--45, 2022.

\bibitem{mehta2024ensemble}
R.~Mehta, A.~Sheikh-Akbari, and K.~K. Singh, ``Ensemble-based hybrid transfer
  approach for an effective 2d ear recognition system,'' \emph{IEEE Access},
  vol.~12, pp. 155\,733--46, 2024.

\bibitem{alomari2024ear}
E.~A.~M. Alomari, S.~Yang, S.~Hoque, and F.~Deravi, ``Ear-based person
  recognition using pix2pix gan augmentation,'' in \emph{Int. Conf. of the
  Biometrics Special Interest Group (BIOSIG)}, 2024, pp. 1--6.

\bibitem{korichi2022tr}
A.~Korichi, S.~Slatnia, and O.~Aiadi, ``Tr-icanet: a fast unsupervised
  deep-learning-based scheme for unconstrained ear recognition,'' \emph{Arabian
  J. for Sci. and Eng.}, vol.~47, no.~8, pp. 9887--98, 2022.

\bibitem{aiadi2023mdfnet}
O.~Aiadi, B.~Khaldi, and C.~Saadeddine, ``Mdfnet: an unsupervised lightweight
  network for ear print recognition,'' \emph{J. of Ambient Intell. and Humaniz.
  Comput.}, vol.~14, no.~10, pp. 13\,773--86, 2023.

\bibitem{he2024self}
J.~He, Y.~He, L.~Zhai, and Y.~Bi, ``Self-supervised siamese networks with
  squeeze-excitation attention for ear image recognition,'' in \emph{Int. Conf.
  on Intell. Comput.}, 2024, pp. 122--33.

\bibitem{freire2025synthesizing}
D.~Freire-Obreg{\'o}n, J.~Neves, {\v{Z}}.~Emer{\v{s}}i{\v{c}}, B.~Meden,
  M.~Castrill{\'o}n-Santana, and H.~Proen{\c{c}}a, ``Synthesizing multilevel
  abstraction ear sketches for enhanced biometric recognition,'' \emph{Image
  and Vis. Comput.}, vol. 154, p. 105424, 2025.

\bibitem{emervsivc2017unconstrained}
{\v{Z}}.~Emer{\v{s}}i{\v{c}}, D.~{\v{S}}tepec, V.~{\v{S}}truc, P.~Peer,
  A.~George, A.~Ahmad, E.~Omar, T.~E. Boult, R.~Safdaii, Y.~Zhou \emph{et~al.},
  ``The unconstrained ear recognition challenge,'' in \emph{Int. Jt. Conf. on
  Biometrics (IJCB)}, 2017, pp. 715--24.

\bibitem{dodge2018unconstrained}
S.~Dodge, J.~Mounsef, and L.~Karam, ``Unconstrained ear recognition using deep
  neural networks,'' \emph{IET Biometrics}, vol.~7, no.~3, pp. 207--14, 2018.

\end{thebibliography}

\onecolumn
\appendices

\setcounter{figure}{0}
\renewcommand{\thefigure}{A.\arabic{figure}}

\section{Recognition Results on Additional Datasets}
\subsection{Identification – CMC Curves Without Fine-Tuning}

This subsection presents the Cumulative Match Characteristic (CMC) curves for the AWE, IITD-II, AMI, and KinEar datasets under the generalization setting (without fine-tuning). These results demonstrate the baseline identification capability of the proposed method across different datasets.

\begin{figure}[!ht]
    \centering
    \begin{subfigure}[b]{0.48\linewidth}
        \includegraphics[width=\linewidth]{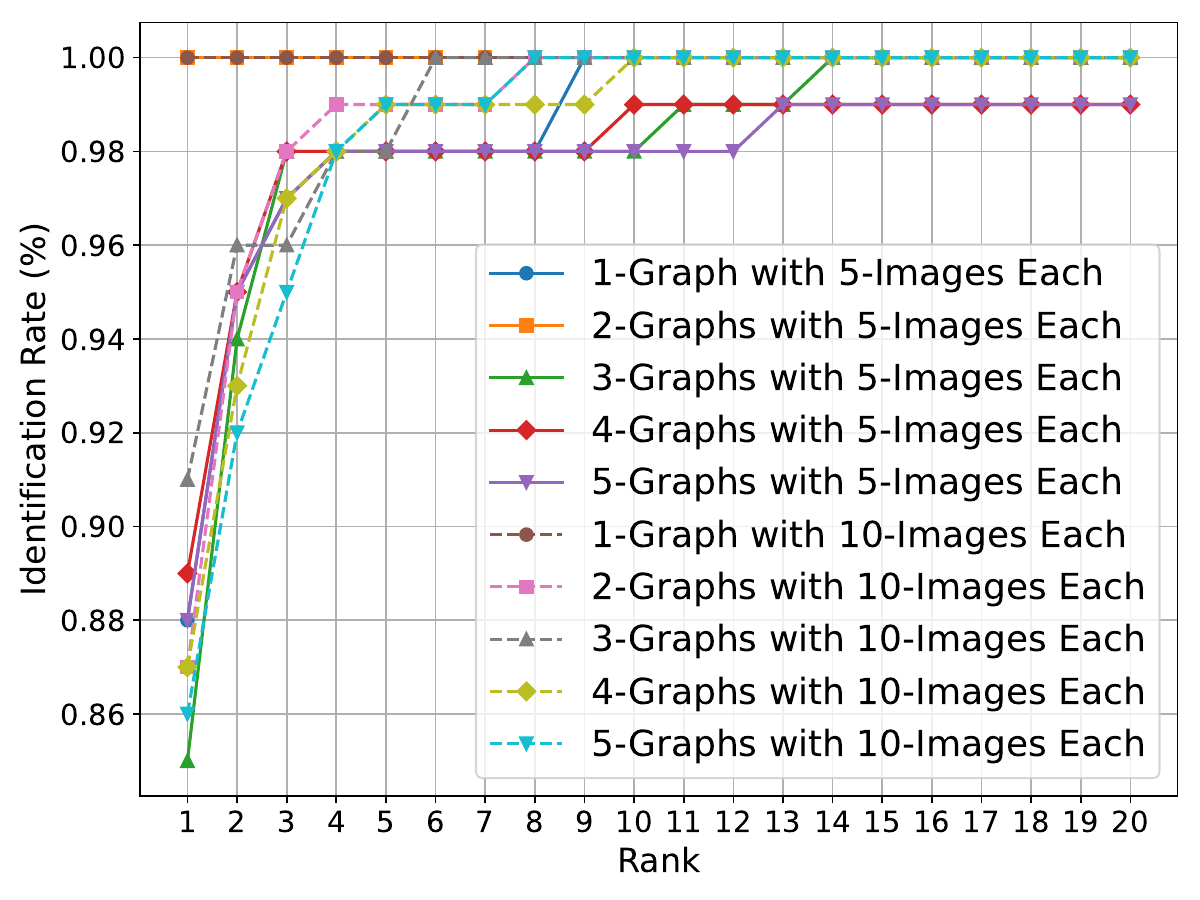}
        \caption{AWE Dataset}
    \end{subfigure}
    \hfill
    \begin{subfigure}[b]{0.48\linewidth}
        \includegraphics[width=\linewidth]{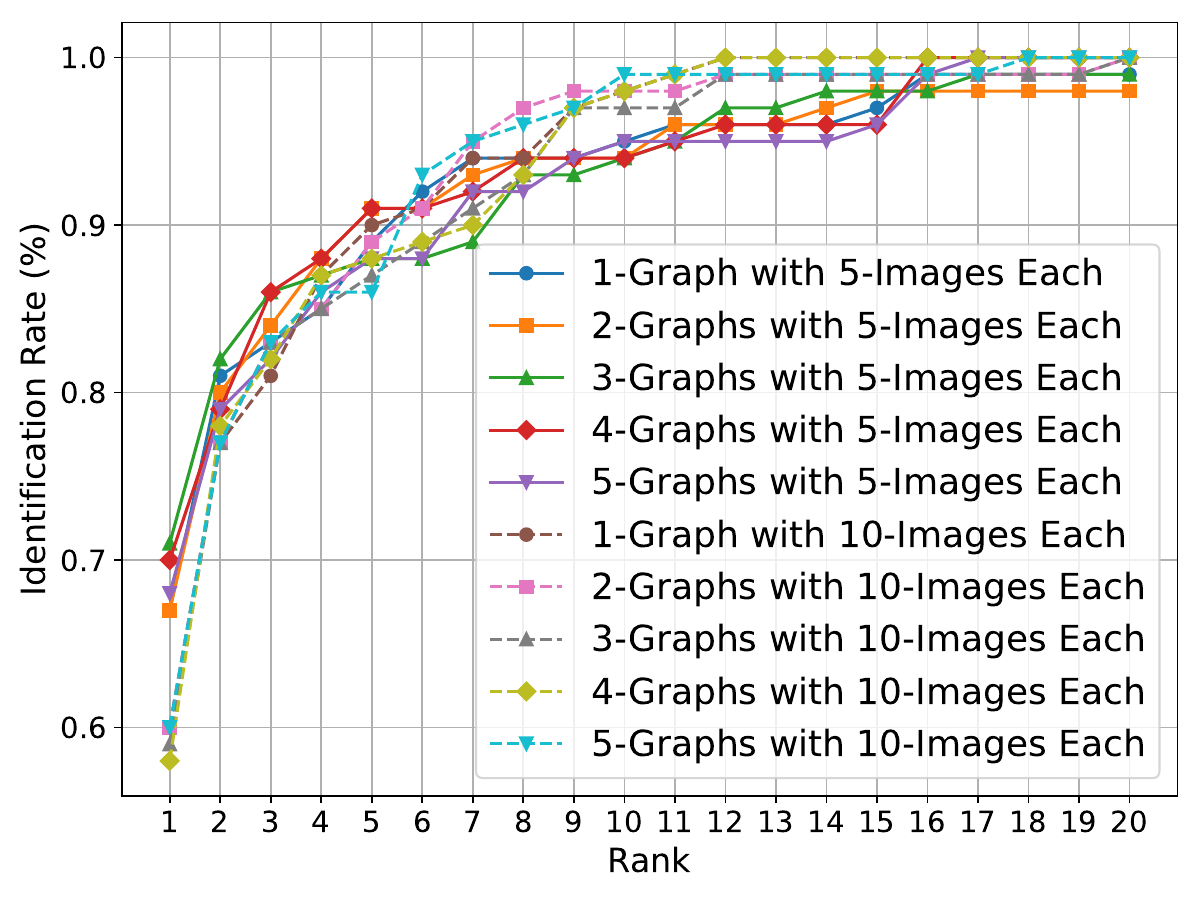}
        \caption{IITD-II Dataset}
    \end{subfigure}
    \vspace{1em}

    \begin{subfigure}[b]{0.48\linewidth}
        \includegraphics[width=\linewidth]{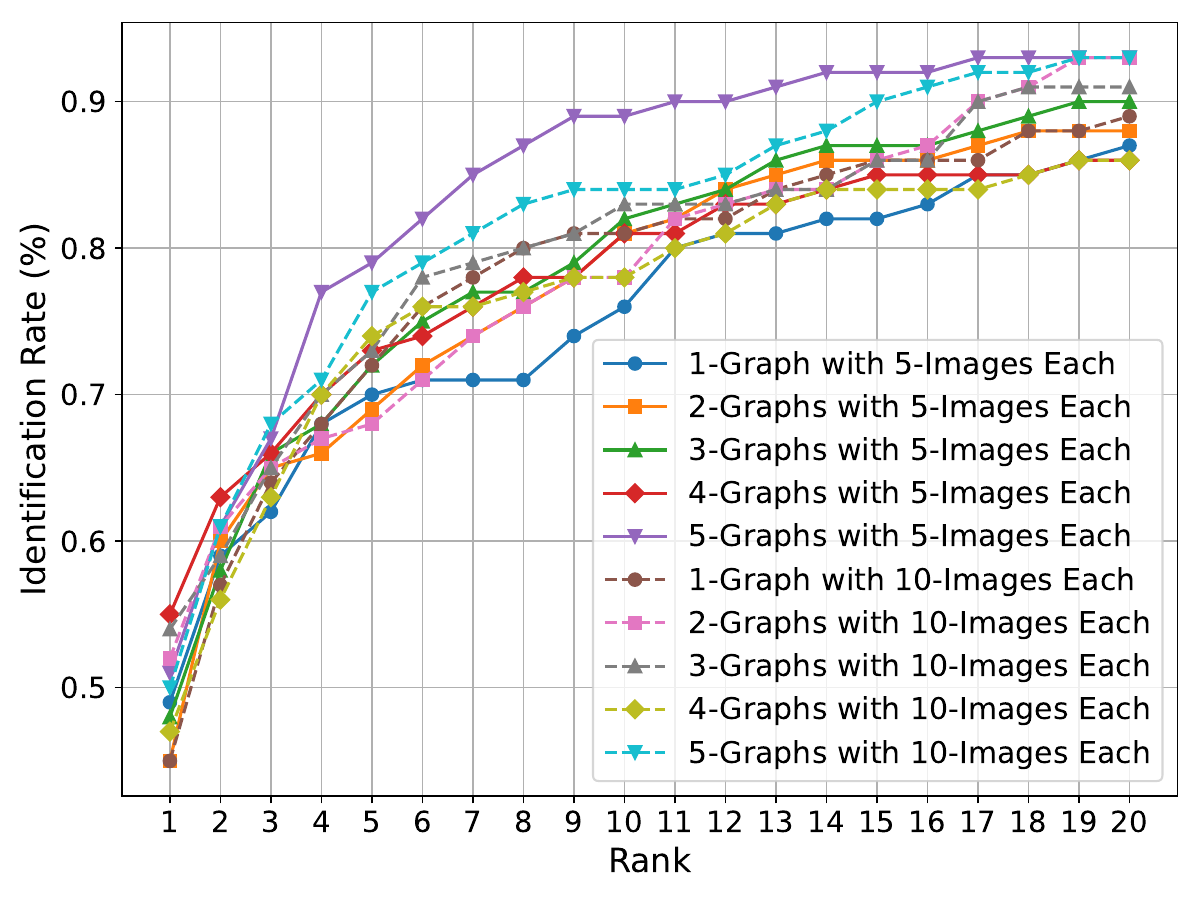}
        \caption{AMI Dataset}
    \end{subfigure}
    \hfill
    \begin{subfigure}[b]{0.48\linewidth}
        \includegraphics[width=\linewidth]{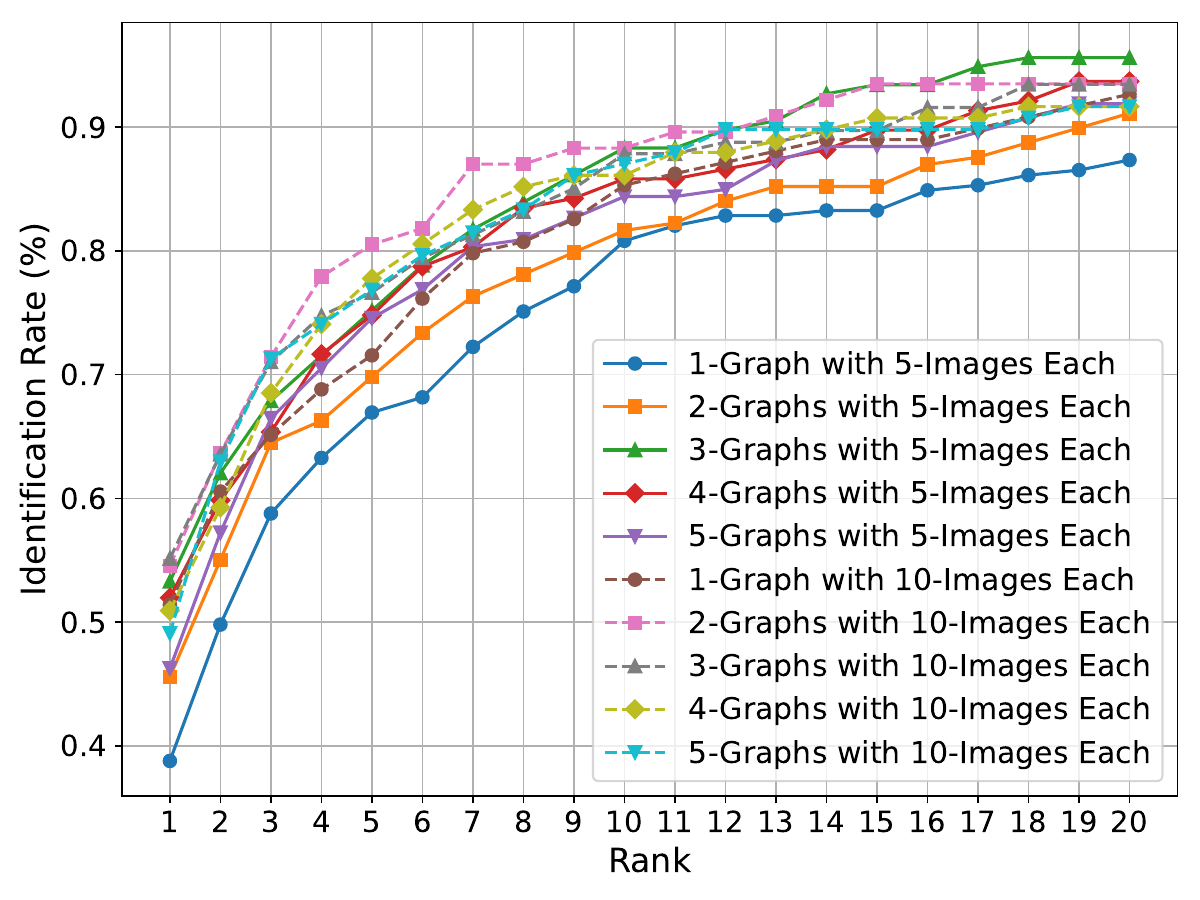}
        \caption{KinEar Dataset}
    \end{subfigure}
    \caption{CMC curves for identification on additional datasets without fine-tuning.}
    \label{fig:s1}
\end{figure}

\newpage
\subsection{Identification – CMC Curves With Fine-Tuning}

This subsection displays CMC curves after fine-tuning the model on the respective datasets, highlighting the improvement in identification performance due to domain-specific adaptation.

\begin{figure}[!ht]
    \centering
    \begin{subfigure}[b]{0.48\linewidth}
        \includegraphics[width=\linewidth]{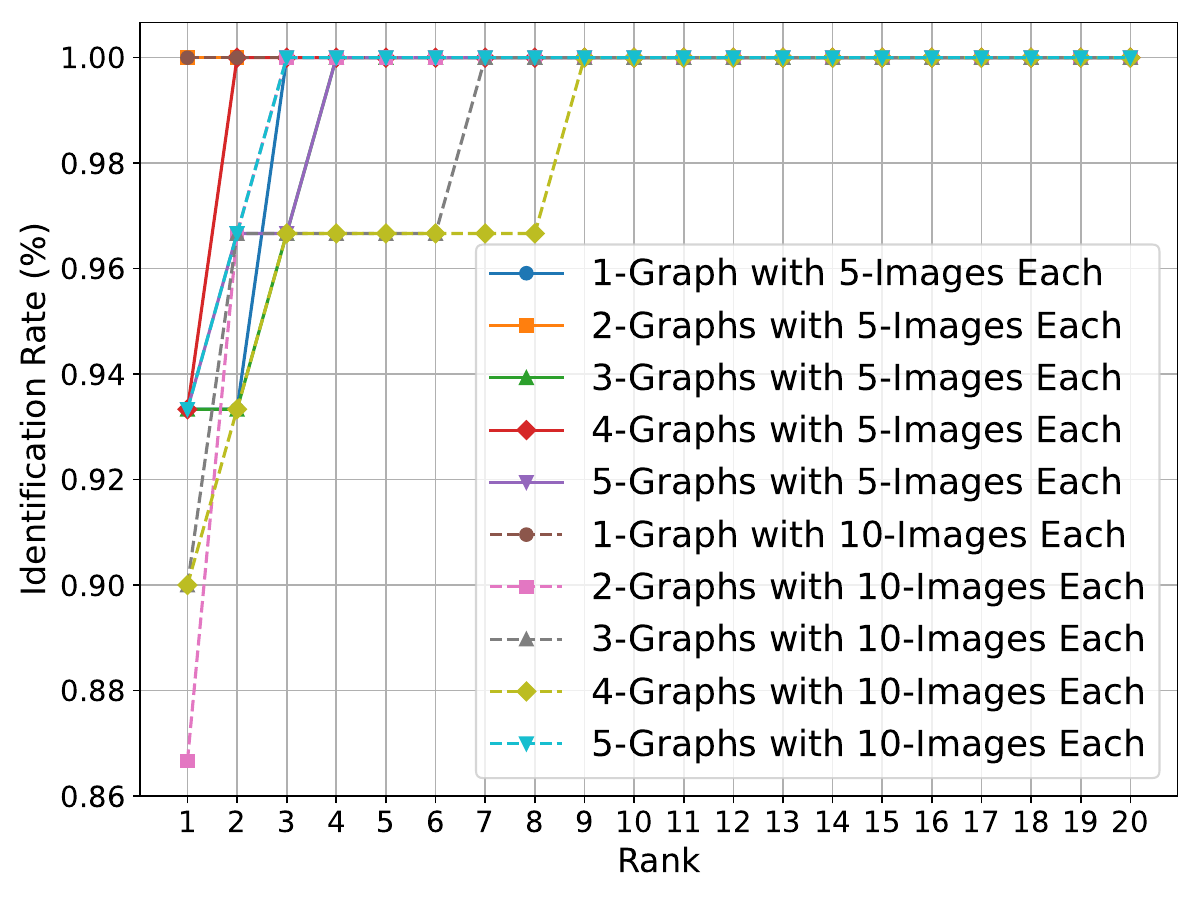}
        \caption{AWE Dataset}
    \end{subfigure}
    \hfill
    \begin{subfigure}[b]{0.48\linewidth}
        \includegraphics[width=\linewidth]{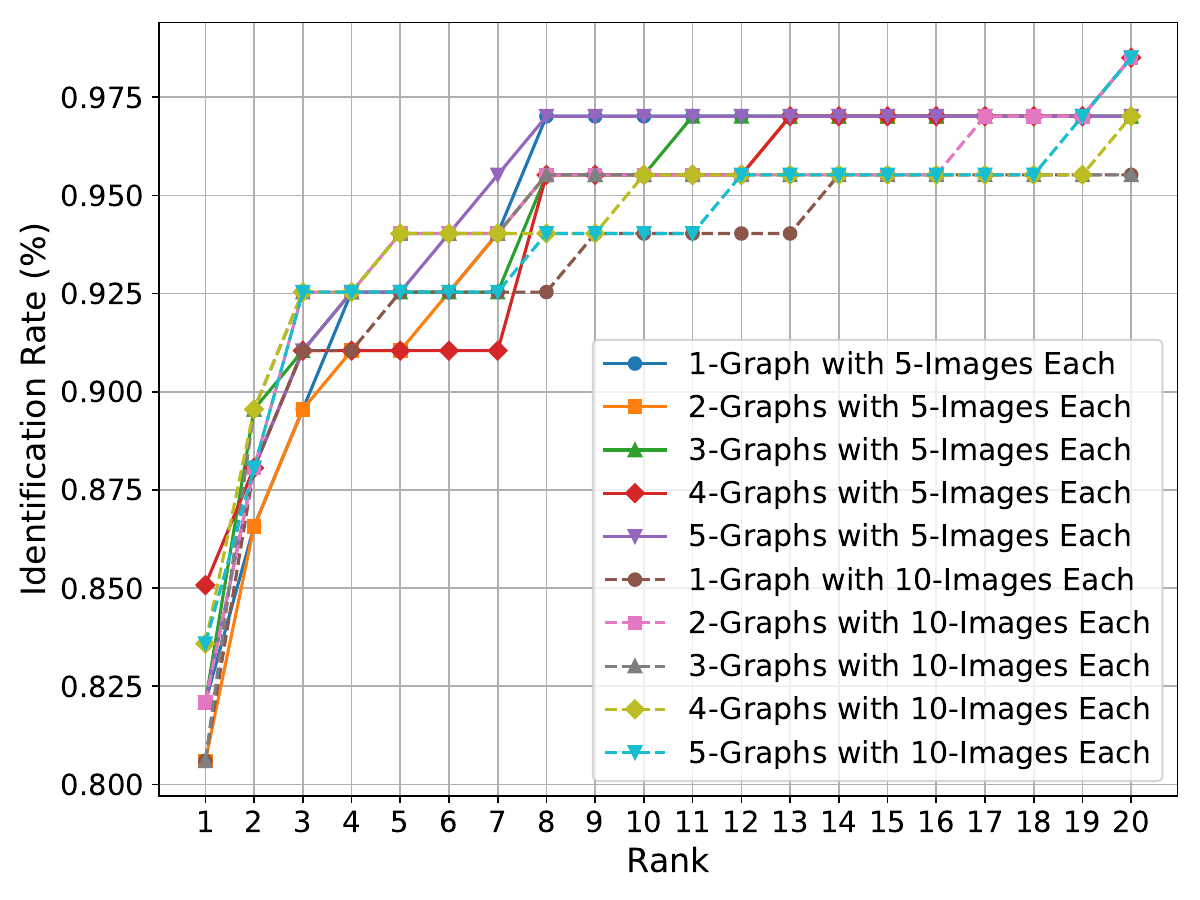}
        \caption{IITD-II Dataset}
    \end{subfigure}
    \vspace{1em}

    \begin{subfigure}[b]{0.48\linewidth}
        \includegraphics[width=\linewidth]{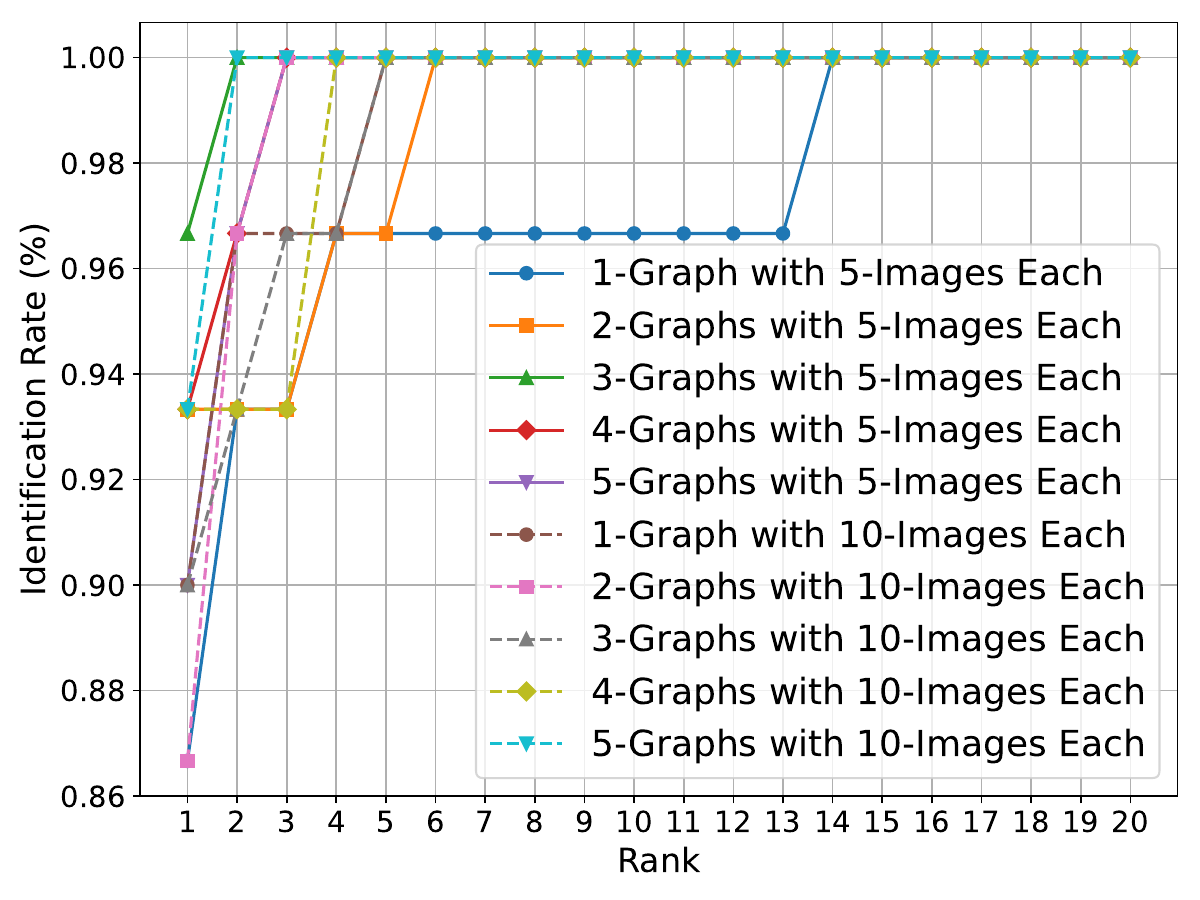}
        \caption{AMI Dataset}
    \end{subfigure}
    \hfill
    \begin{subfigure}[b]{0.48\linewidth}
        \includegraphics[width=\linewidth]{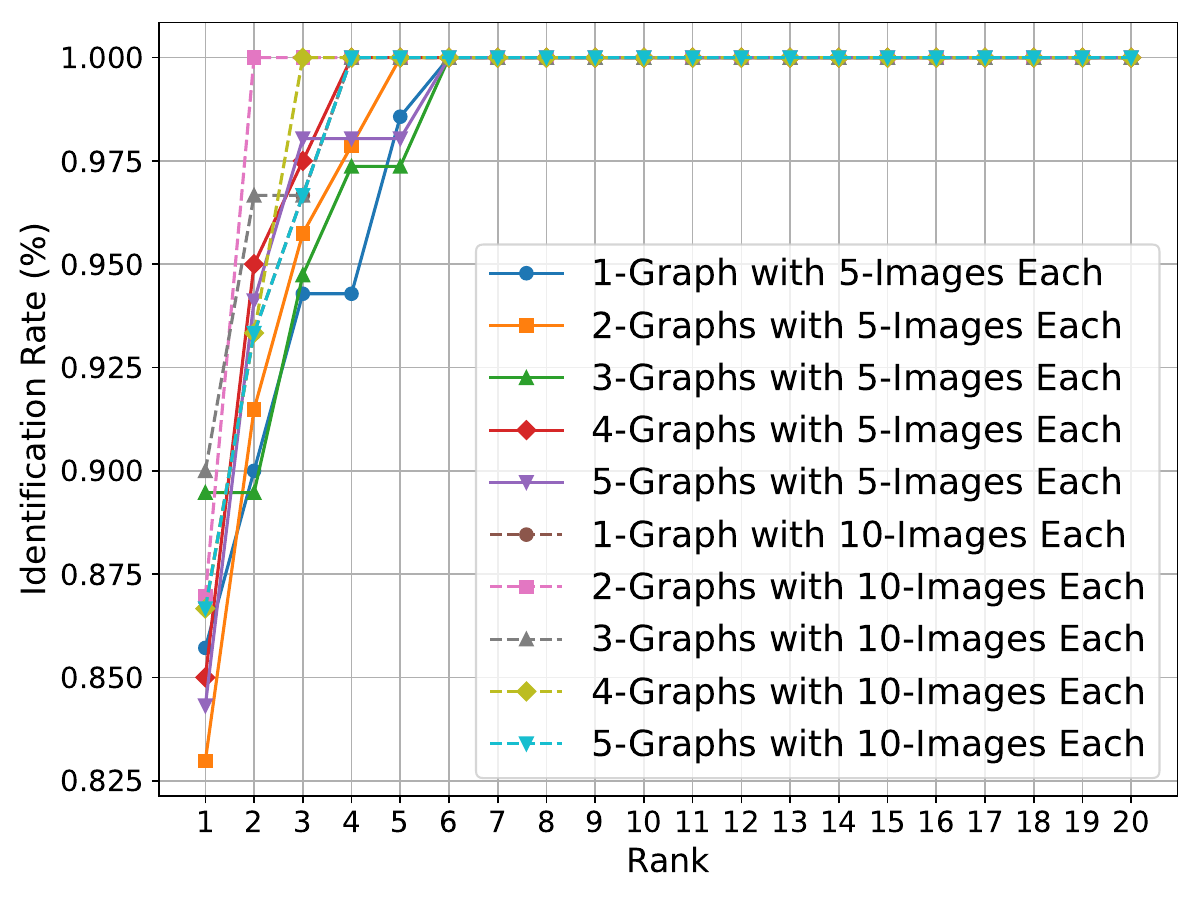}
        \caption{KinEar Dataset}
    \end{subfigure}
    \caption{CMC curves for identification on additional datasets with fine-tuning.}
    \label{fig:s2}
\end{figure}

\newpage
\subsection{Verification – ROC Curves Without Fine-Tuning}

This subsection reports ROC curves on additional datasets without fine-tuning. These results reflect the model’s generalization ability for the verification task under unseen domain settings.

\begin{figure}[!ht]
    \centering
    \begin{subfigure}[b]{0.48\linewidth}
        \includegraphics[width=\linewidth]{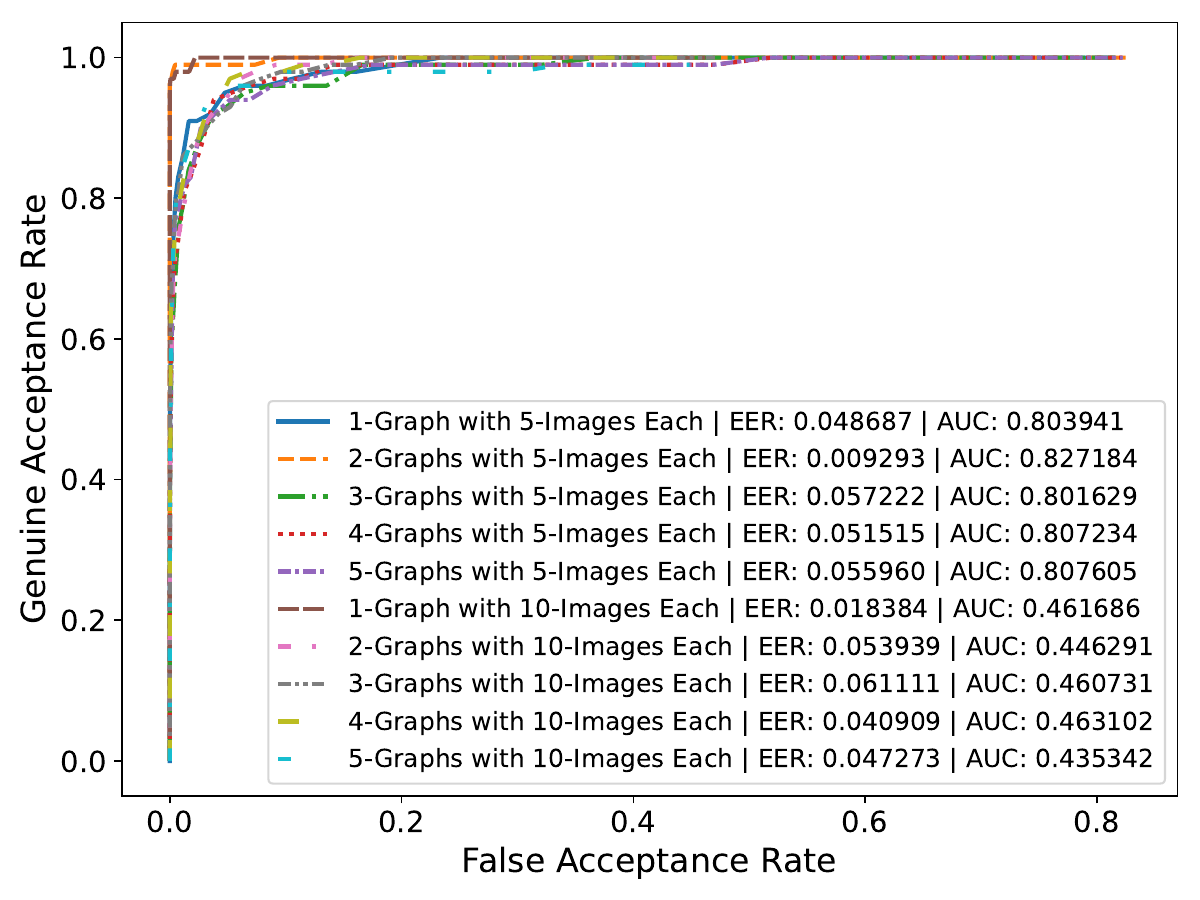}
        \caption{AWE Dataset}
    \end{subfigure}
    \hfill
    \begin{subfigure}[b]{0.48\linewidth}
        \includegraphics[width=\linewidth]{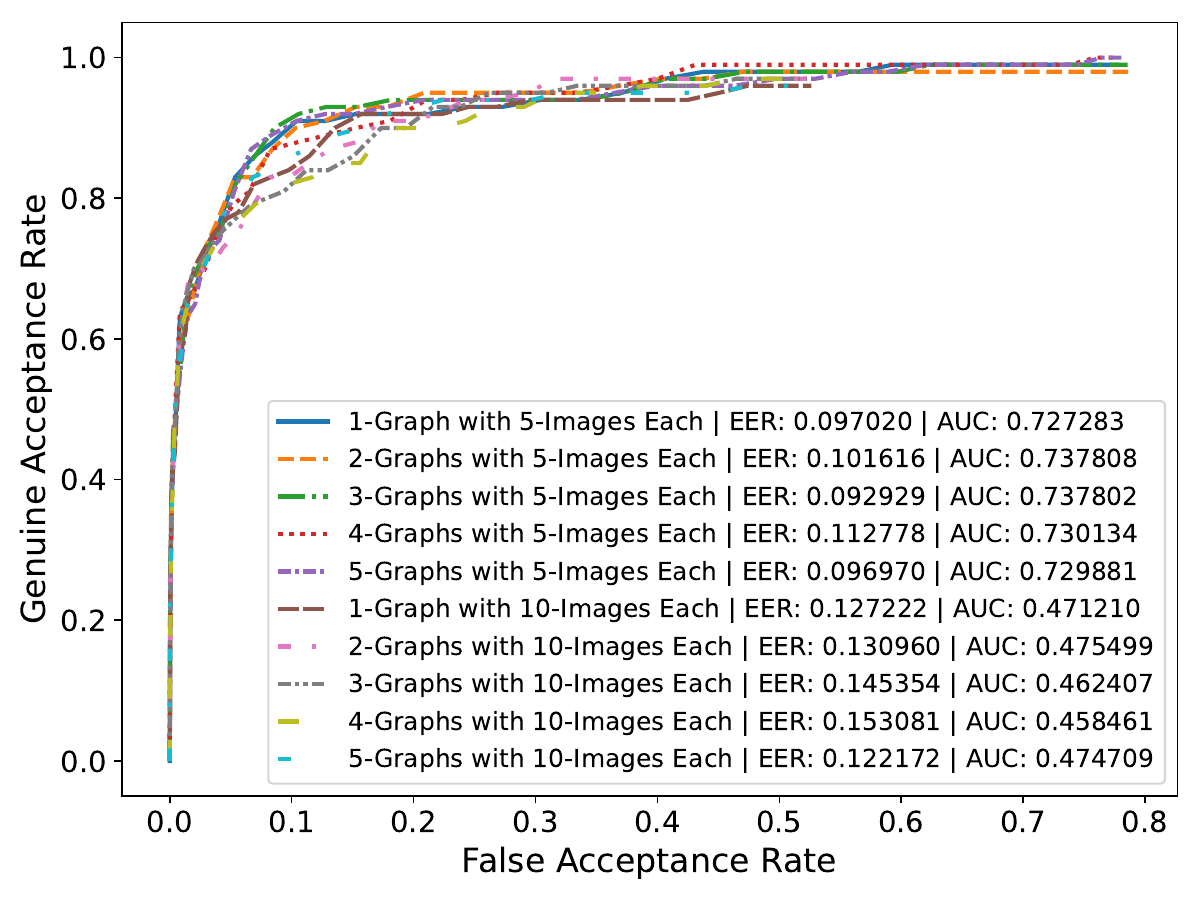}
        \caption{IITD-II Dataset}
    \end{subfigure}
    \vspace{1em}

    \begin{subfigure}[b]{0.48\linewidth}
        \includegraphics[width=\linewidth]{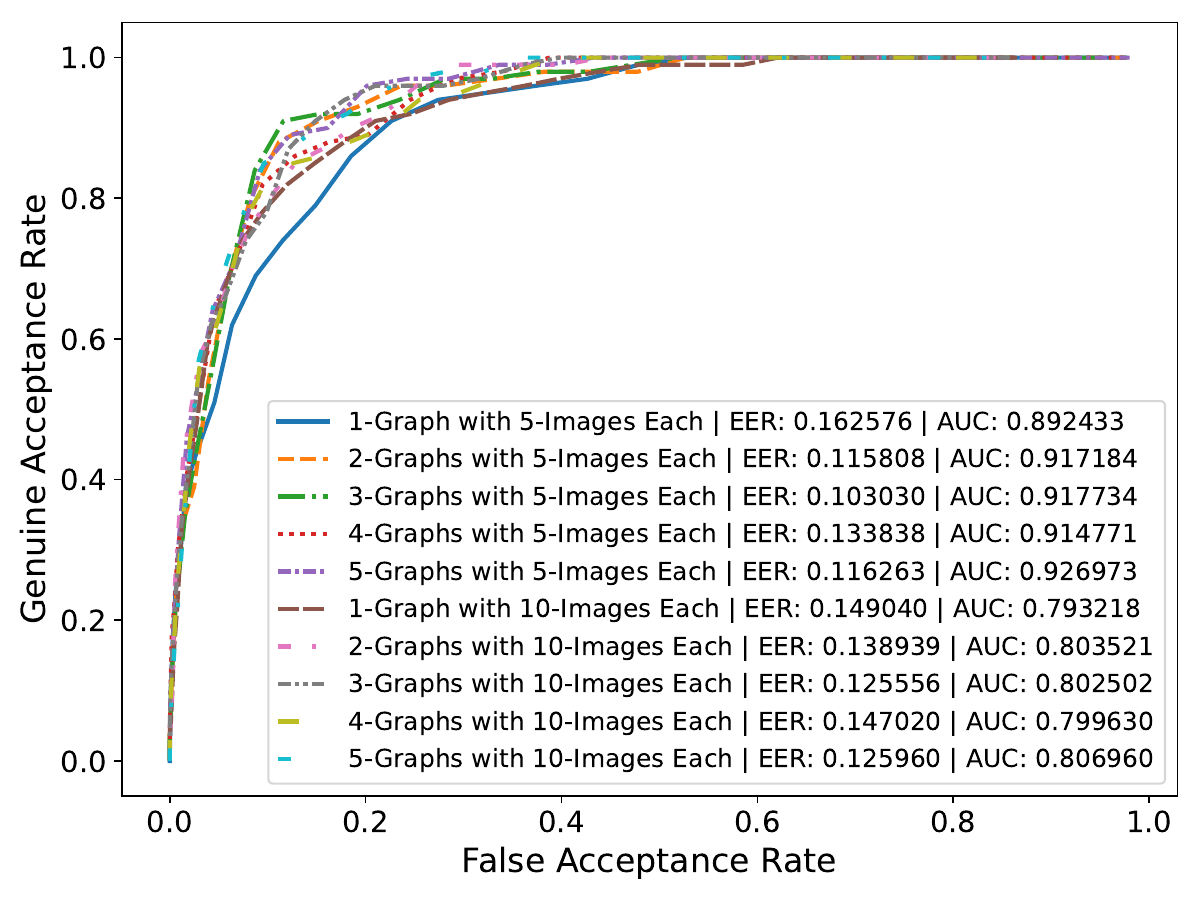}
        \caption{AMI Dataset}
    \end{subfigure}
    \hfill
    \begin{subfigure}[b]{0.48\linewidth}
        \includegraphics[width=\linewidth]{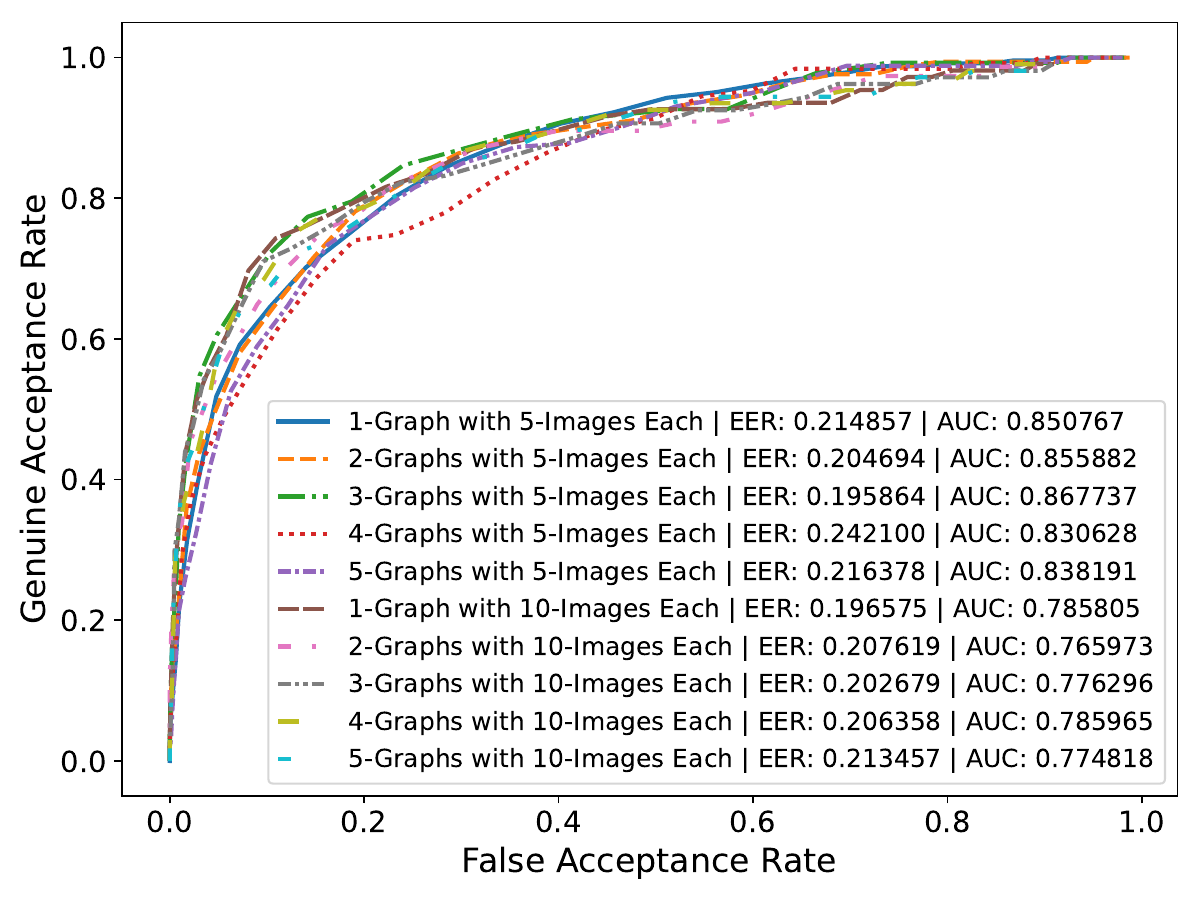}
        \caption{KinEar Dataset}
    \end{subfigure}
    \caption{ROC curves for verification on additional datasets without fine-tuning.}
    \label{fig:s3}
\end{figure}

\newpage
\subsection{Verification – ROC Curves With Fine-Tuning}

This subsection shows the ROC curves for verification performance after fine-tuning the model on the respective datasets. It illustrates the effect of domain adaptation on genuine/imposter discrimination.

\begin{figure}[!ht]
    \centering
    \begin{subfigure}[b]{0.48\linewidth}
        \includegraphics[width=\linewidth]{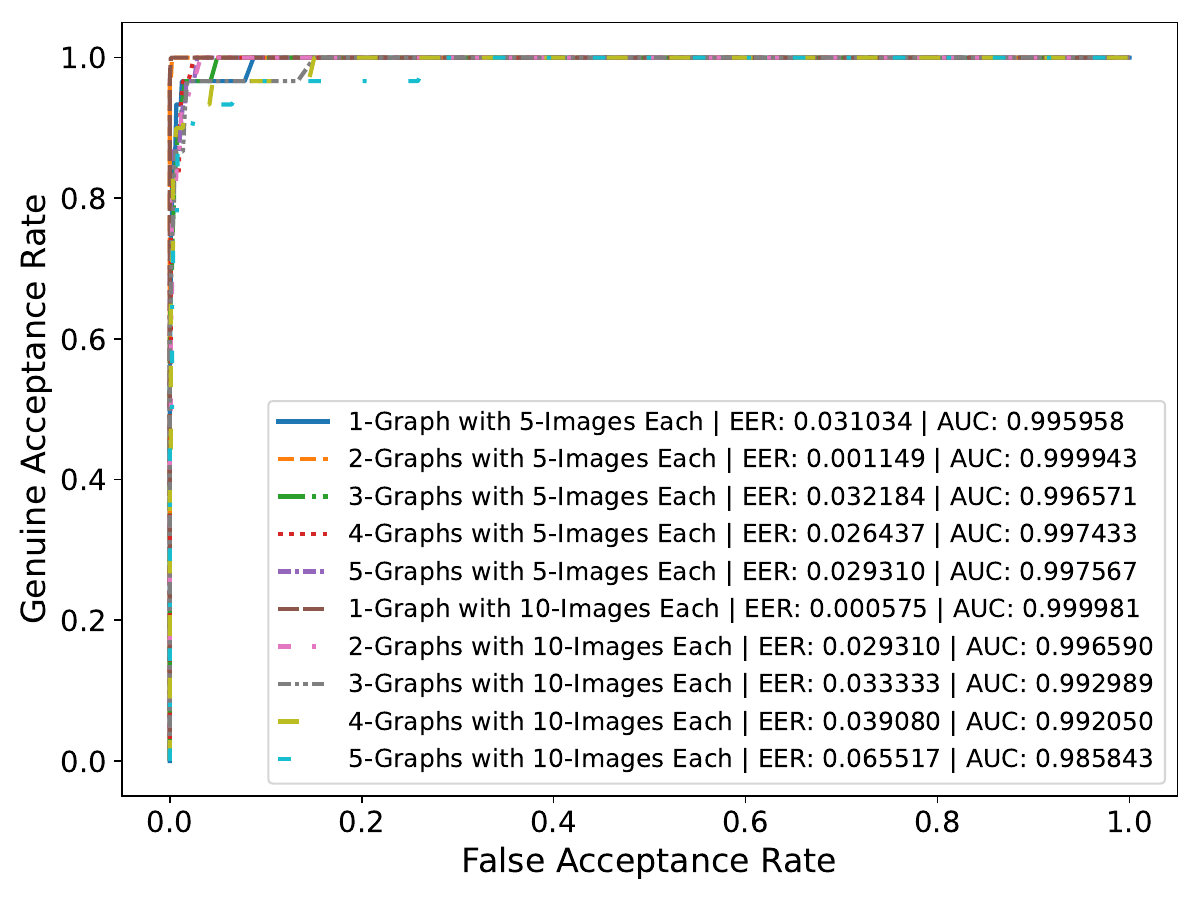}
        \caption{AWE Dataset}
    \end{subfigure}
    \hfill
    \begin{subfigure}[b]{0.48\linewidth}
        \includegraphics[width=\linewidth]{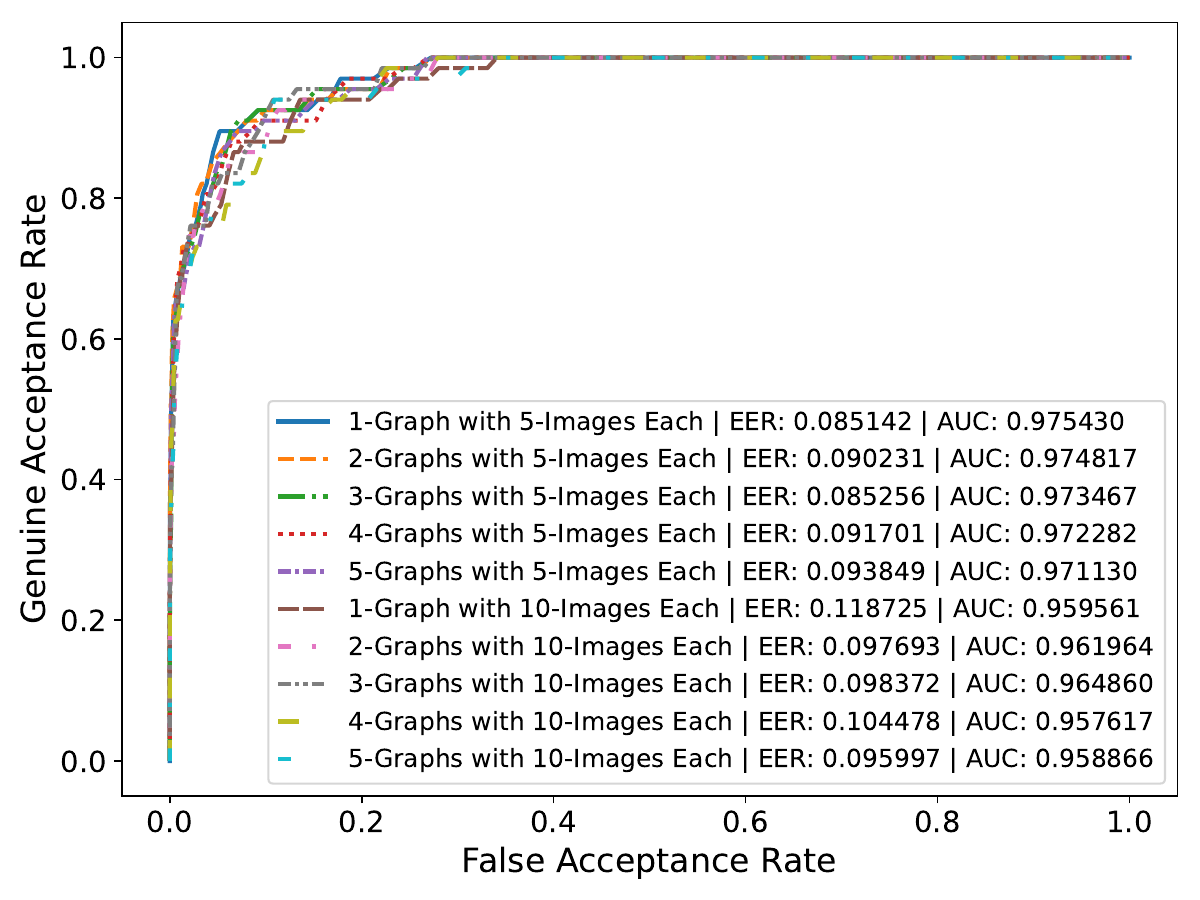}
        \caption{IITD-II Dataset}
    \end{subfigure}
    \vspace{1em}

    \begin{subfigure}[b]{0.48\linewidth}
        \includegraphics[width=\linewidth]{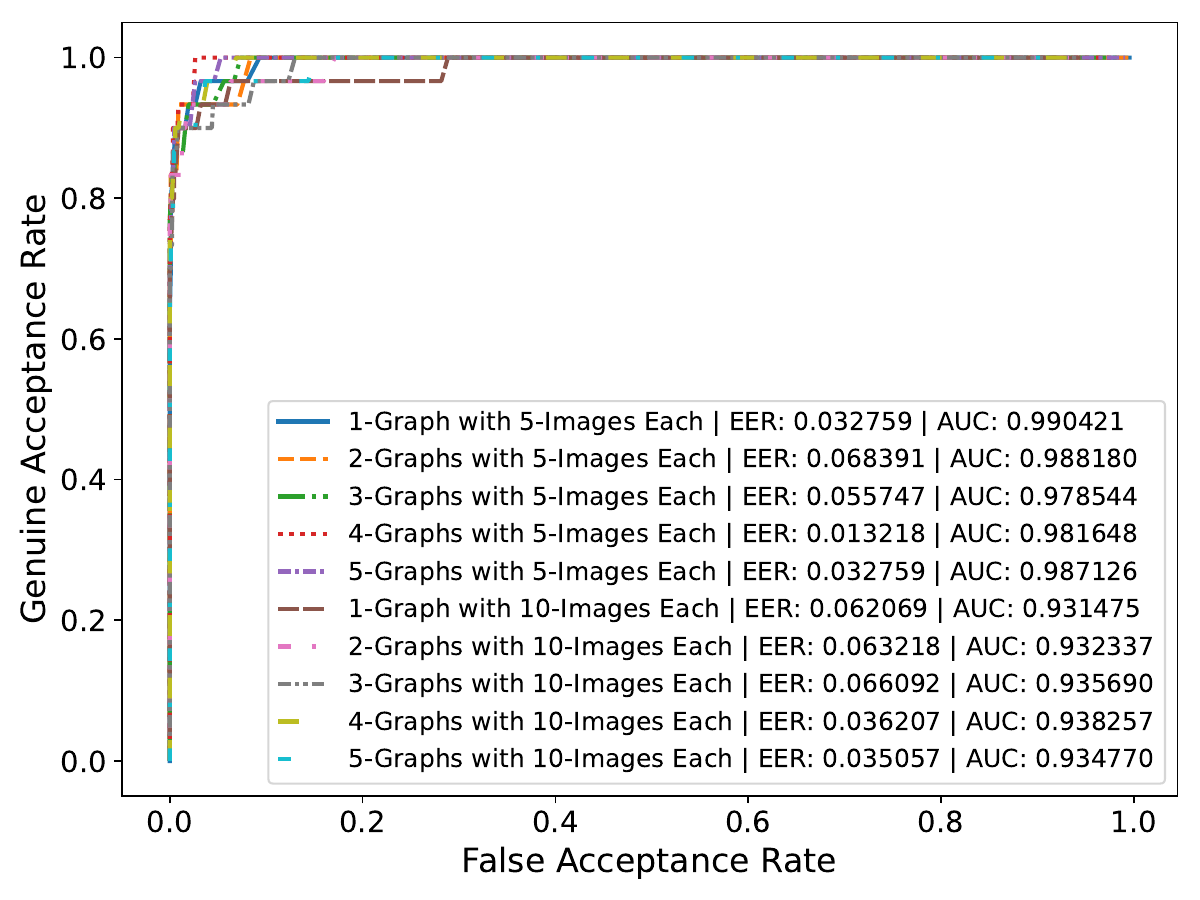}
        \caption{AMI Dataset}
    \end{subfigure}
    \hfill
    \begin{subfigure}[b]{0.48\linewidth}
        \includegraphics[width=\linewidth]{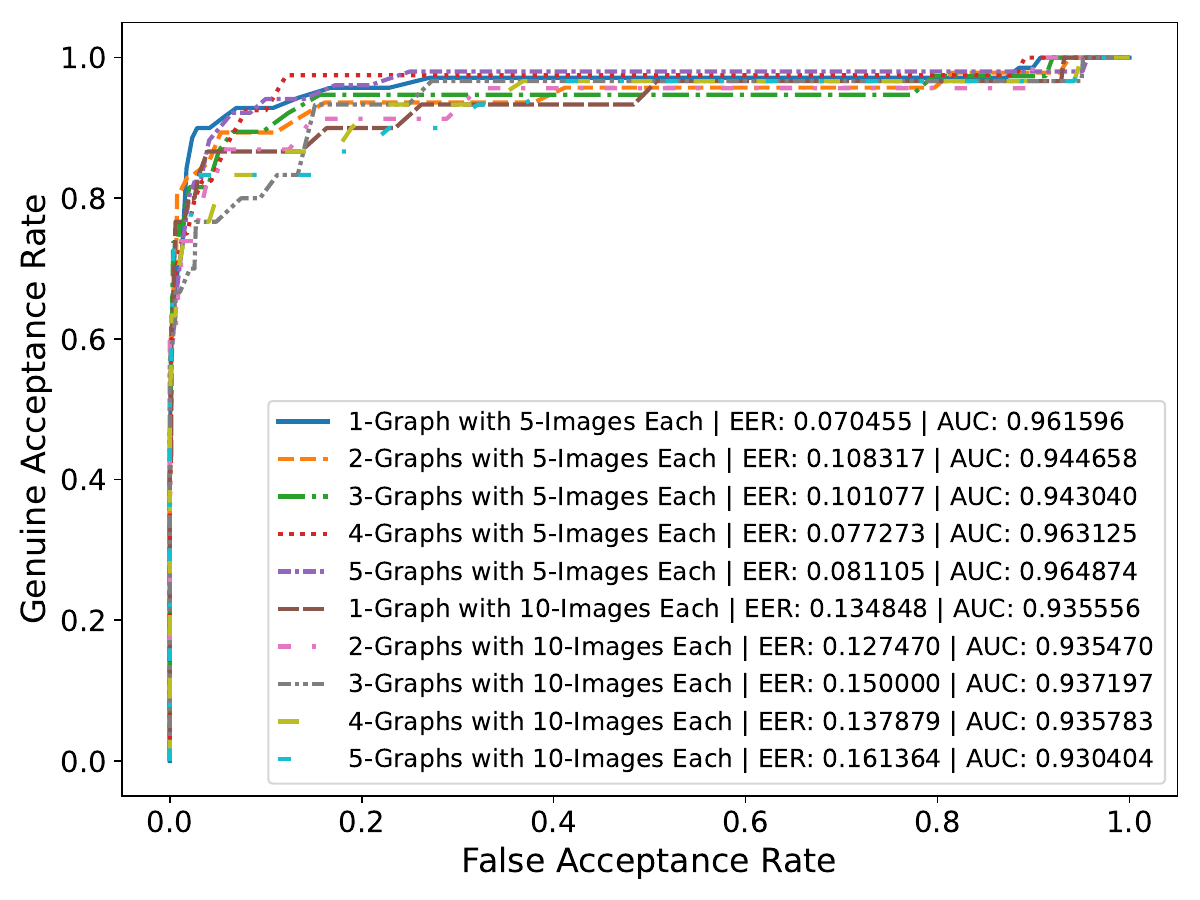}
        \caption{KinEar Dataset}
    \end{subfigure}
    \caption{ROC curves for verification on additional datasets with fine-tuning.}
    \label{fig:s4}
\end{figure}

\newpage
\section{Additional Ablation Study Figures}
This section includes visualizations from ablation studies analyzing the contribution of key components.

\begin{figure*}[!ht]

    \centering
    \begin{subfigure}[b]{0.4\linewidth}
        \includegraphics[width=\linewidth]{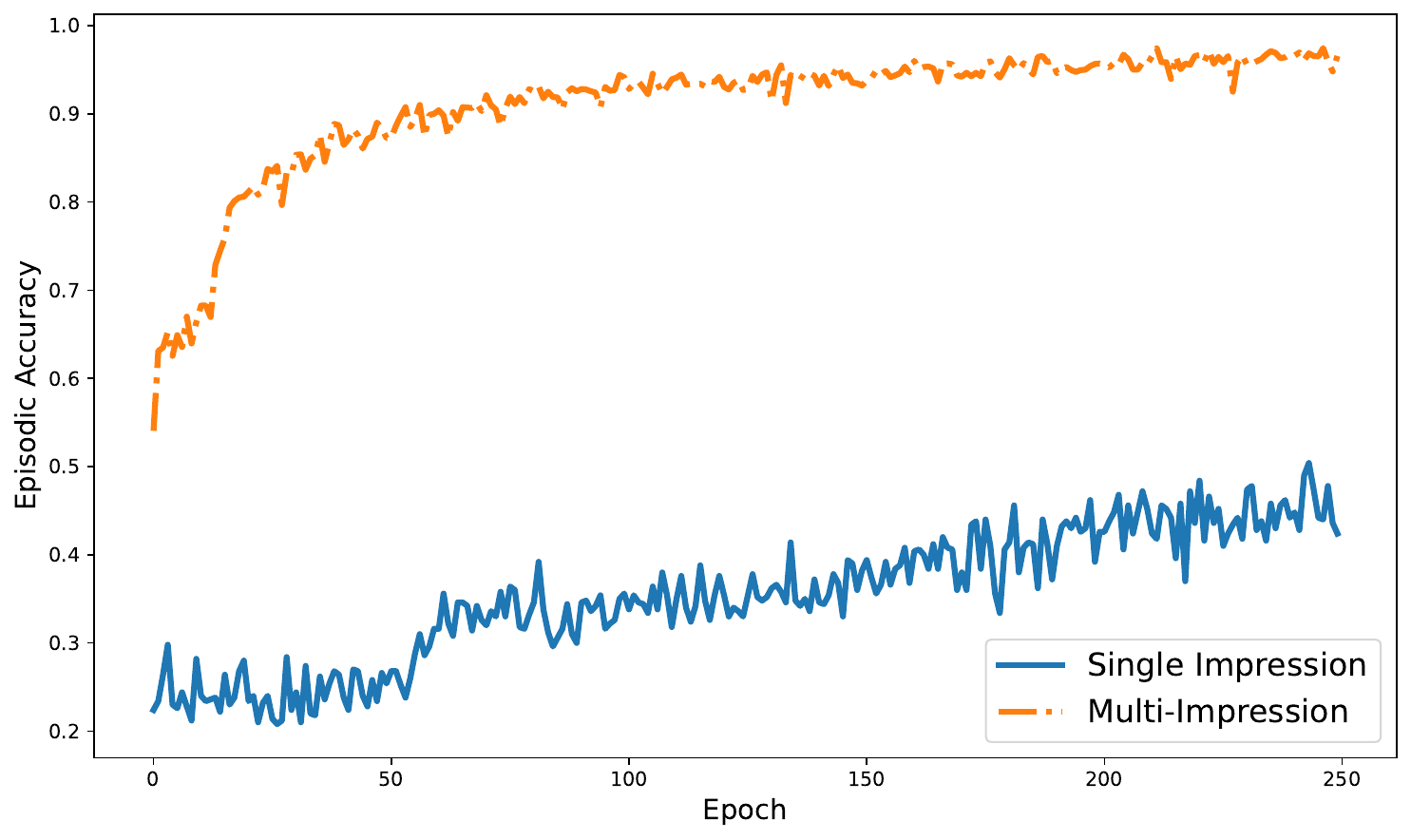}
        \caption{Single vs. Multi Impression (Episodic)}
    \end{subfigure}
    \begin{subfigure}[b]{0.4\linewidth}
        \includegraphics[width=\linewidth]{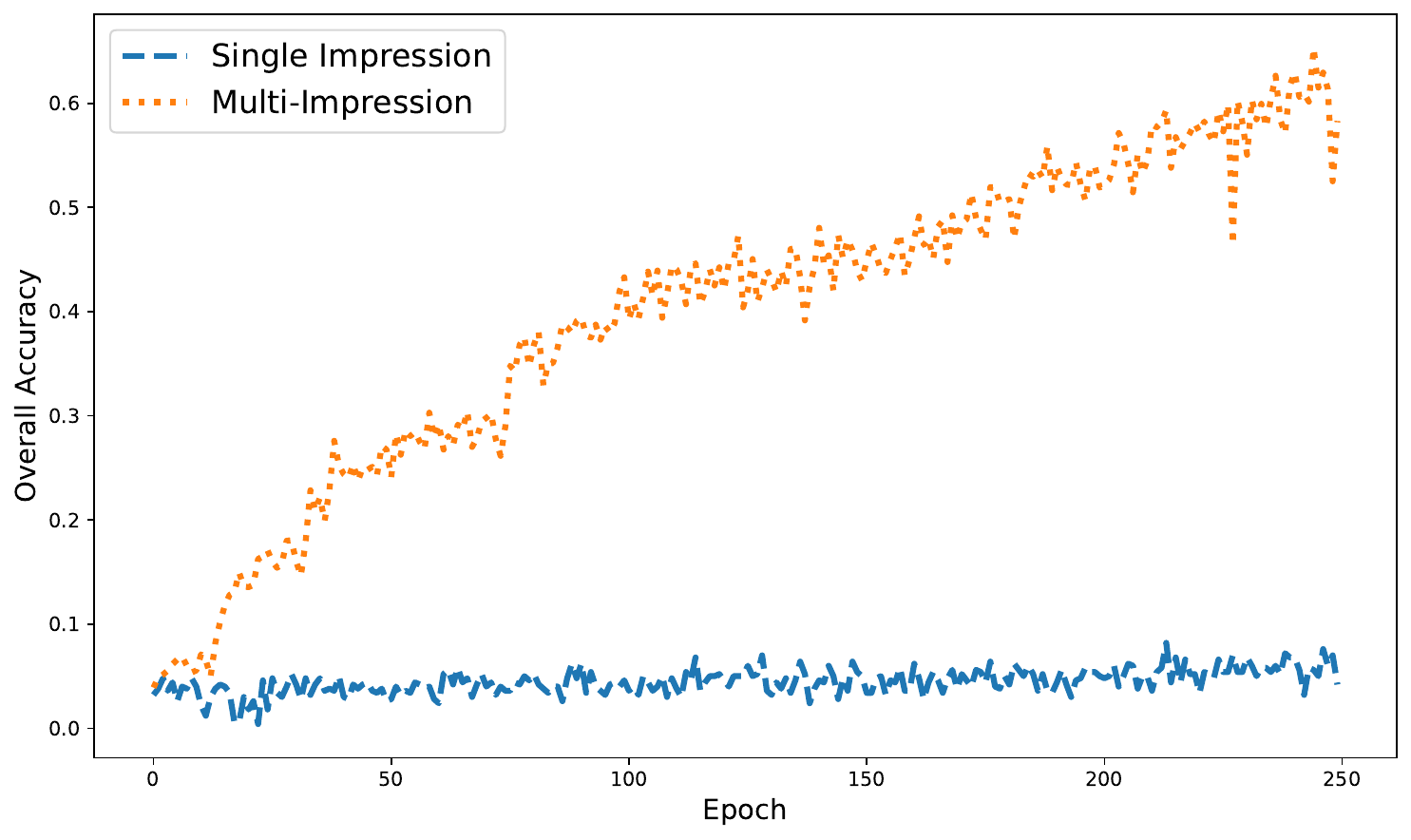}
        \caption{Single vs. Multi Impression (Overall)}
    \end{subfigure}
    \vspace{1em}

    \centering
    \begin{subfigure}[b]{0.4\linewidth}
        \includegraphics[width=\linewidth]{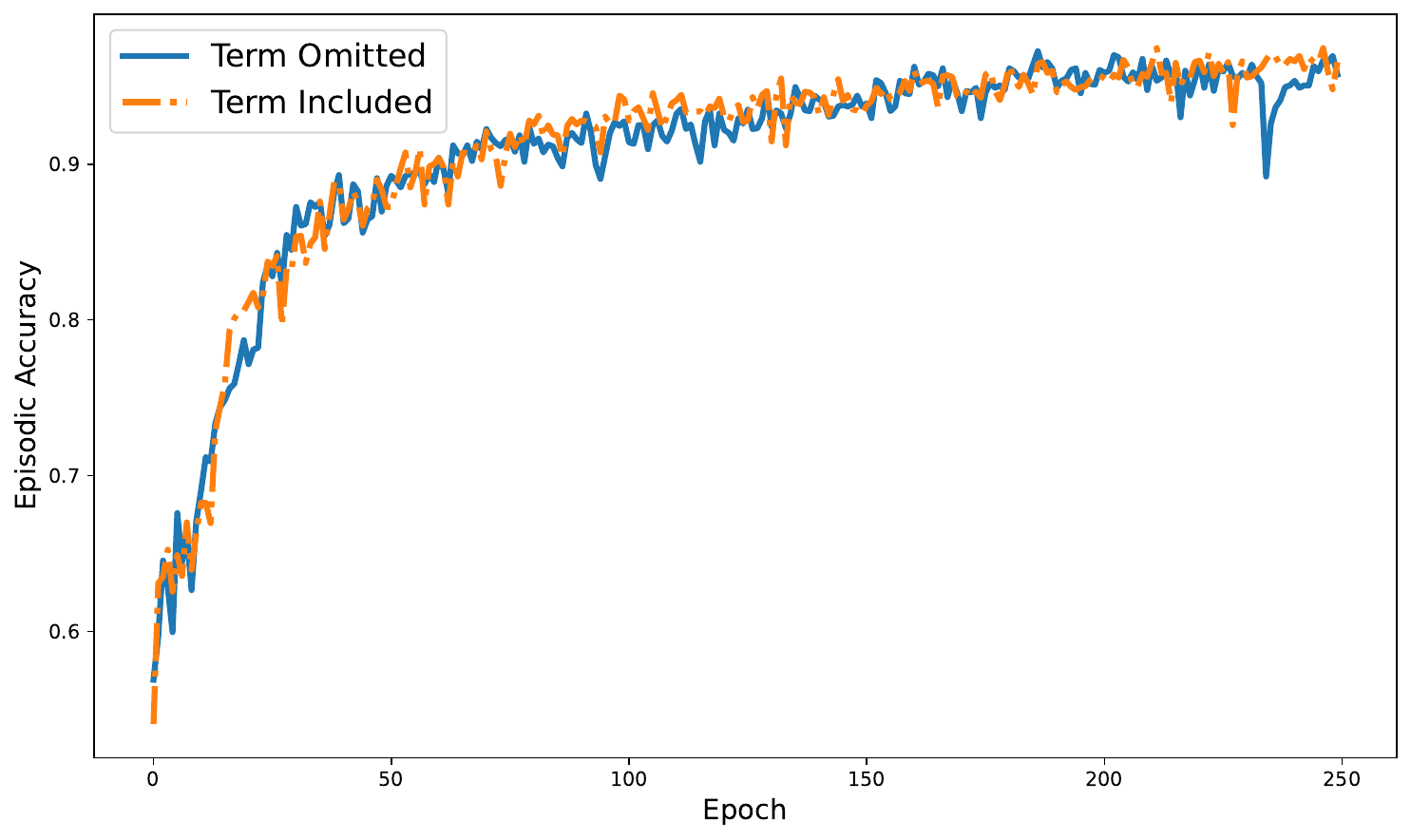}
        \caption{Cross-Graph Alignment Term (Episodic)}
    \end{subfigure}
    \begin{subfigure}[b]{0.4\linewidth}
        \includegraphics[width=\linewidth]{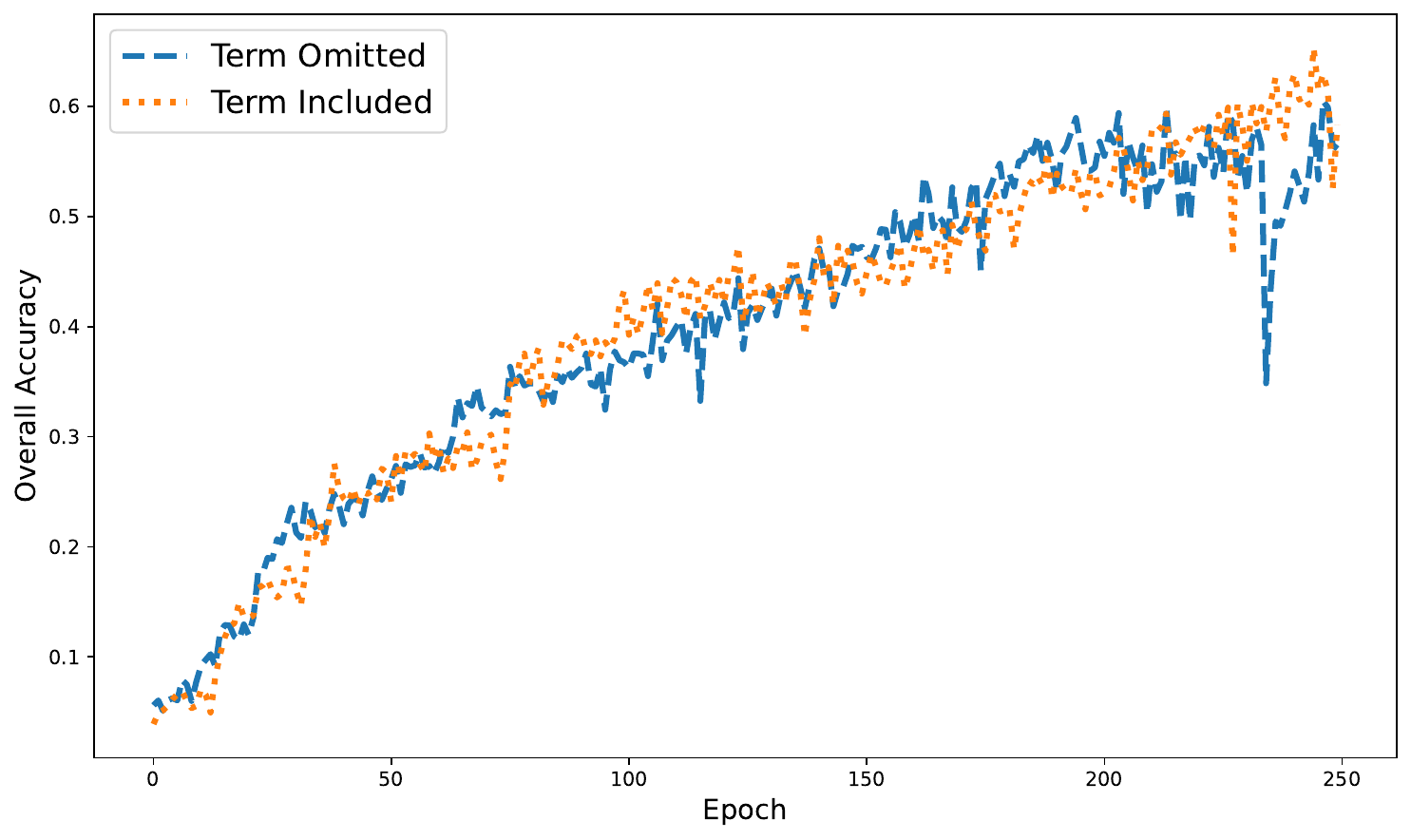}
        \caption{Cross-Graph Alignment Term (Overall)}
    \end{subfigure}
    \vspace{1em}

    \centering
    \begin{subfigure}[b]{0.4\linewidth}
        \includegraphics[width=\linewidth]{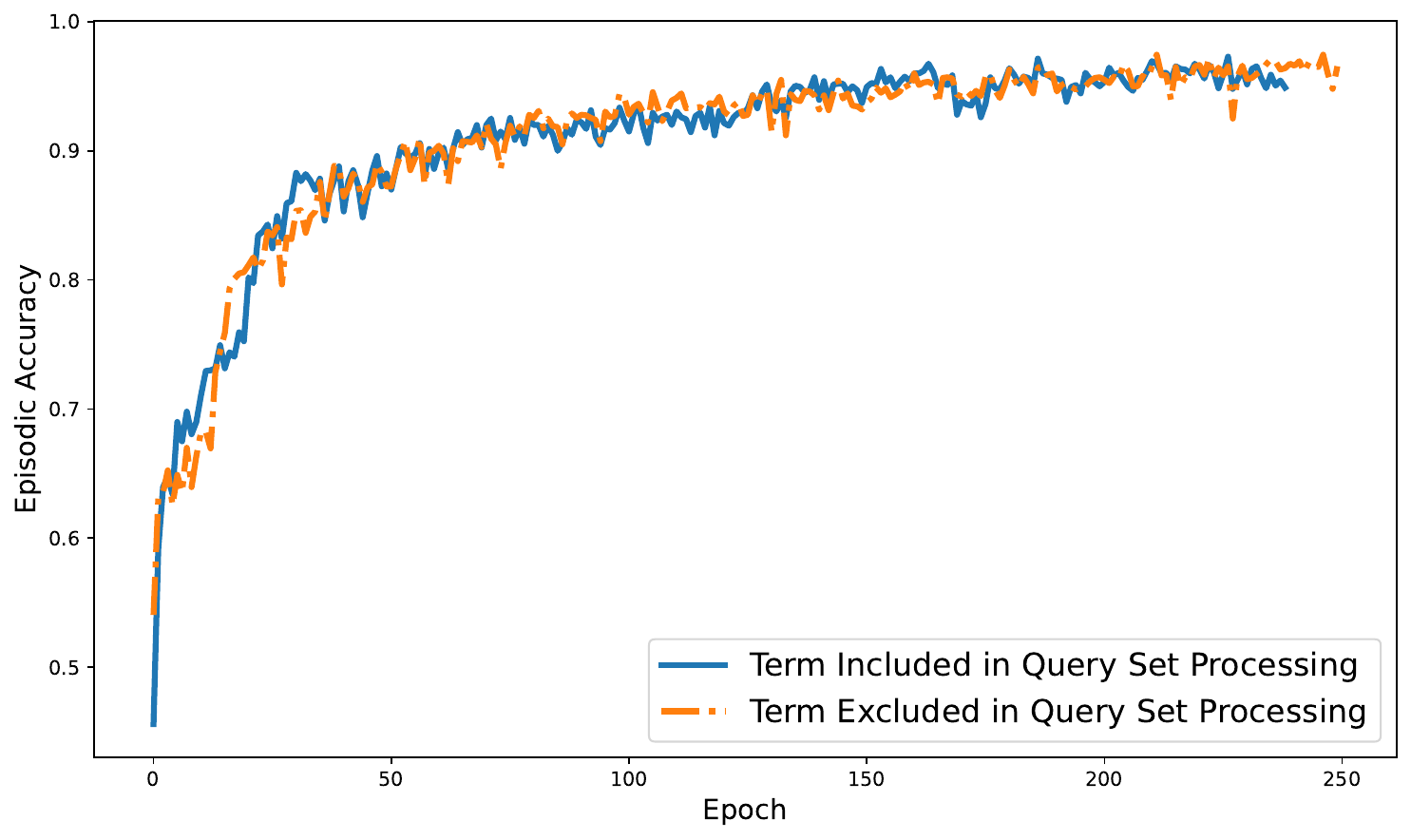}
        \caption{Prototype Alignment Term in Query Graph (Episodic)}
    \end{subfigure}
    \begin{subfigure}[b]{0.4\linewidth}
        \includegraphics[width=\linewidth]{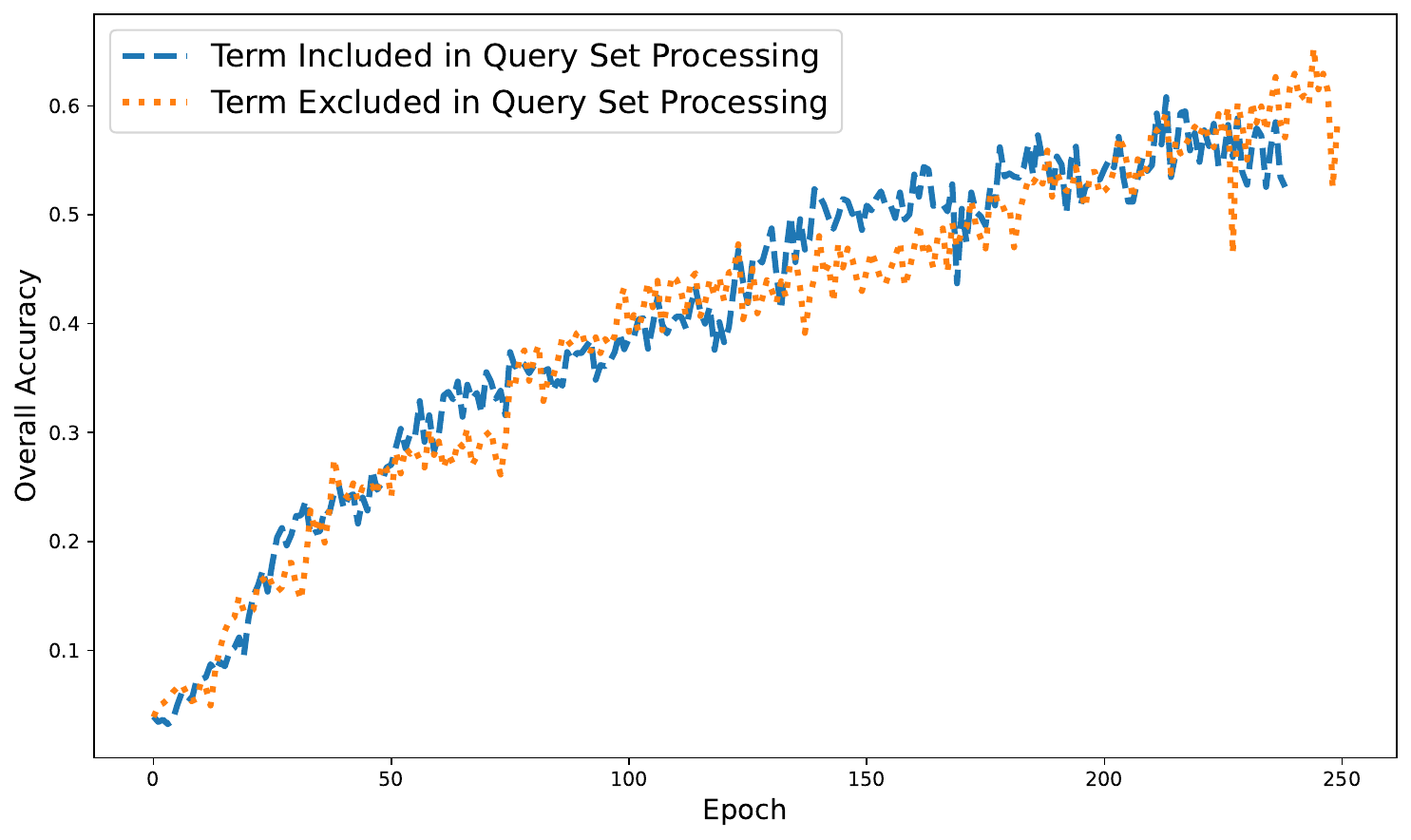}
        \caption{Prototype Alignment Term in Query Graph (Overall)}
    \end{subfigure}
    \vspace{1em}

    \centering
    \begin{subfigure}[b]{0.4\linewidth}
        \includegraphics[width=\linewidth]{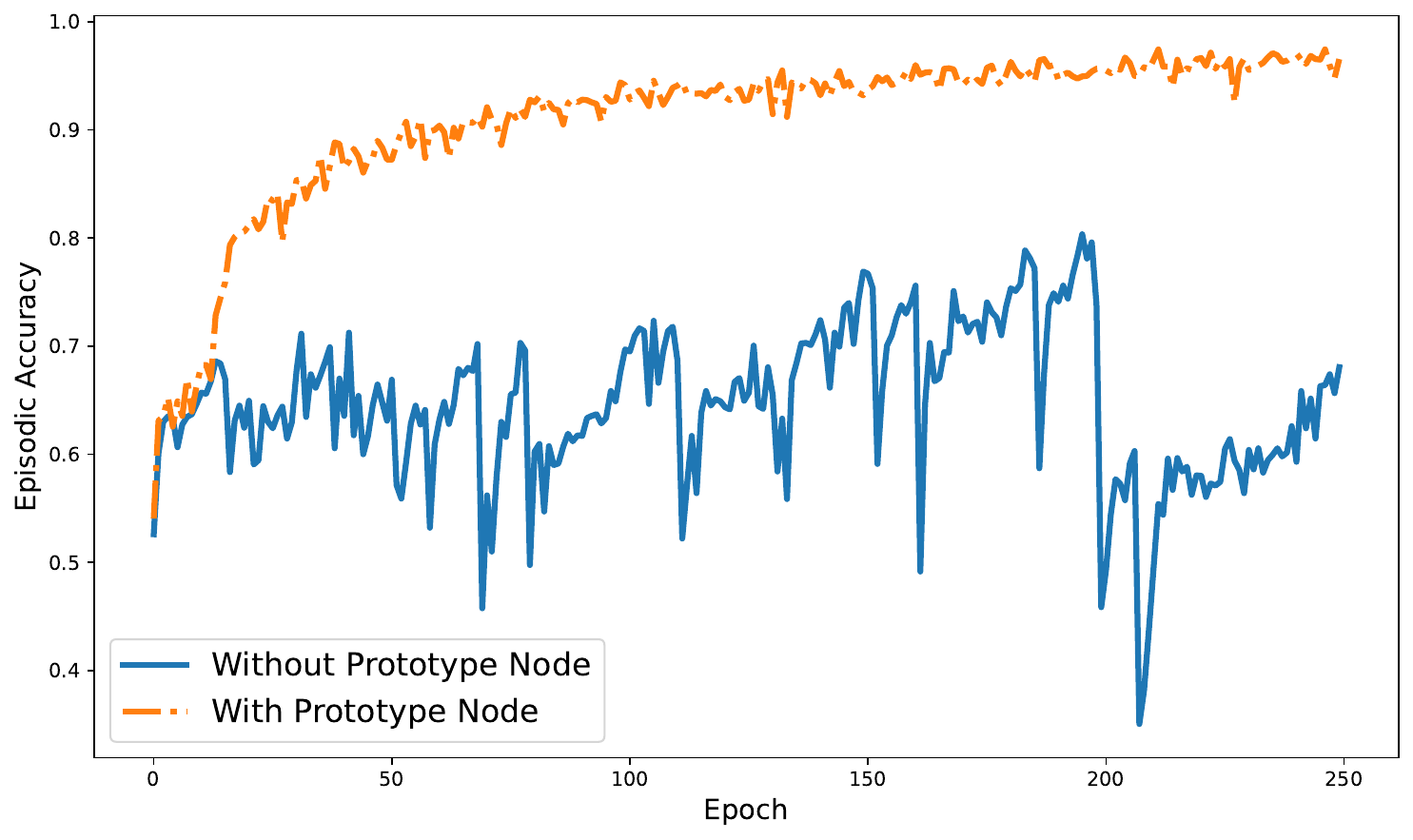}
        \caption{Impact of Prototype Node (Episodic)}
    \end{subfigure}
    \begin{subfigure}[b]{0.4\linewidth}
        \includegraphics[width=\linewidth]{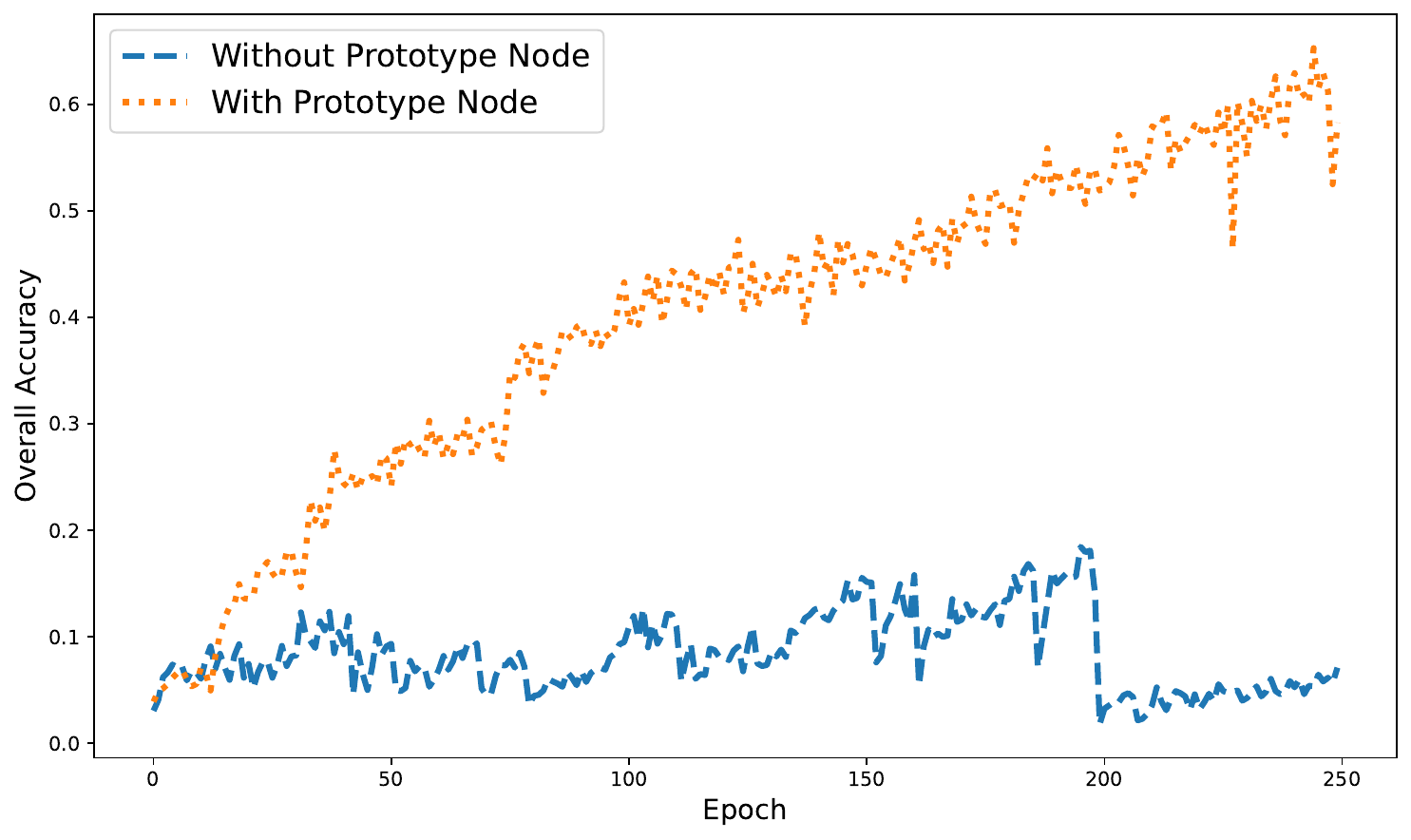}
        \caption{Impact of Prototype Node (Overall)}
    \end{subfigure}  

    \caption{Impact of ablation configurations on model performance.}
    \label{fig:s5}
    
\end{figure*}

\end{document}